\documentclass[10pt,twocolumn,letterpaper]{article}

\usepackage[pagenumbers]{wacv} 

\usepackage{graphicx}
\usepackage{amsmath}
\usepackage{amssymb}
\usepackage{booktabs}
\usepackage{lipsum}
\usepackage{rotating}
\usepackage{appendix}
\usepackage{multirow}
\usepackage{graphicx}
\usepackage[table]{xcolor}

\usepackage{xcolor}
\usepackage{pifont}

\usepackage{float} 
\usepackage{stfloats}
\definecolor{best}{RGB}{128, 179, 255}
\definecolor{2best}{RGB}{200, 240, 255}
\definecolor{cornflowerblue}{rgb}{0.39, 0.58, 0.93}
\usepackage{pgfplots}
\usepackage{pgfplotstable}
\usepackage{algorithm}
\usepackage{algpseudocode}
\usepackage{listings}
\newenvironment{conditions}
  {\par\vspace{\abovedisplayskip}\noindent
   \begin{tabular}{>{$}l<{$} @{} >{${}}c<{{}$} @{} l}}
  {\end{tabular}\par\vspace{\belowdisplayskip}}
  
\usepackage{listings}
\usepackage{xcolor}

\definecolor{codegreen}{rgb}{0,0.6,0}
\definecolor{codegray}{rgb}{0.5,0.5,0.5}
\definecolor{codepurple}{rgb}{0.58,0,0.82}
\definecolor{backcolour}{rgb}{0.95,0.95,0.92}

\lstdefinestyle{mystyle}{
    backgroundcolor=\color{backcolour},   
    commentstyle=\color{codegreen},
    keywordstyle=\color{magenta},
    numberstyle=\tiny\color{codegray},
    stringstyle=\color{codepurple},
    basicstyle=\ttfamily\footnotesize,
    breakatwhitespace=false,         
    breaklines=true,                 
    captionpos=b,                    
    keepspaces=true,                 
    numbers=left,                    
    numbersep=5pt,                  
    showspaces=false,                
    showstringspaces=false,
    showtabs=false,                  
    tabsize=2
}

\definecolor{commentcolor}{RGB}{110,154,155}   
\usepackage{pgfplots}
\usepackage{tikz}
\pgfplotsset{compat=1.18}
\usetikzlibrary{pgfplots.fillbetween}
\usepgfplotslibrary{groupplots}
\usetikzlibrary{shapes, positioning, arrows.meta}

\usepackage{booktabs} 
\usepackage{array} 

\usepackage{graphicx}
\usepackage{subcaption}
\usepackage{tikz}

\definecolor{wacvblue}{rgb}{0.21,0.49,0.74}
\usepackage[pagebackref,breaklinks,colorlinks,allcolors=wacvblue]{hyperref}

\def\wacvPaperID{1911} 
\def\confName{WACV}
\def\confYear{2026}

\title{Latent Uncertainty Representations for Video-based Driver Action and Intention Recognition}

\author{
    Koen Vellenga\textsuperscript{\tiny{1,2}} \and
    H. Joe Steinhauer\textsuperscript{1} \and
    Jonas Andersson\textsuperscript{\tiny{2}} \and
    Anders Sjögren\textsuperscript{\tiny{2}} \\
        \textsuperscript{\tiny{1}}{\small University of Skövde, Sweden}, 
    \textsuperscript{\tiny{2}}{\small Volvo Car Corporation, Sweden}
}

\begin{document}
\maketitle

\begin{abstract}
Deep neural networks (DNNs) are increasingly applied to safety-critical tasks in resource-constrained environments, such as video-based driver action and intention recognition. While last layer probabilistic deep learning (LL--PDL) methods can detect out-of-distribution (OOD) instances, their performance varies. As an alternative to last layer approaches, we propose extending pre-trained DNNs with transformation layers to produce multiple latent representations to estimate the uncertainty. We evaluate our latent uncertainty representation (LUR) and repulsively trained LUR (RLUR) approaches against eight PDL methods across four video-based driver action and intention recognition datasets, comparing classification performance, calibration, and uncertainty-based OOD detection. We also contribute 28,000 frame-level action labels and 1,194 video-level intention labels for the NuScenes dataset. Our results show that LUR and RLUR achieve comparable in-distribution classification performance to other LL--PDL approaches. For uncertainty-based OOD detection, LUR matches top-performing PDL methods while being more efficient to train and easier to tune than approaches that require Markov-Chain Monte Carlo sampling or repulsive training procedures. Code and annotations: \url{https://github.com/koenvellenga/LUR}.
\end{abstract}

\section{Introduction}
\label{sec:intro}
Advanced driver assistance systems (ADAS) or autonomous driving systems (ADS) will inevitably encounter novel and complex traffic scenarios that were not represented in the training datasets. In the scenario where human and (semi) autonomous vehicles coexist on the road, it is crucial for such systems to understand human behavior (\eg, the actions or intentions of drivers). Recognizing driving behavior is complex because driving maneuvers can overlap, are influenced by cultural norms and driving context, and may not always be rational \cite{gabriel2020artificial}. Given that human behavior can be irrational and the safety-critical operating environment, it is vital to asses what scenarios a system can confidently predict. However, state-of-the-art (SOTA) video-based driver action and intention recognition approaches commonly rely on deterministic deep neural networks (DNNs) \cite{yang2023aide,noguchi2023ego,ma2023cemformer,vellenga2024evaluation} that do not express uncertainty about the produced predictions. Moreover, DNNs are opaque and difficult to interpret, poorly calibrated \cite{guo2017calibration}, overconfident \cite{hein2019relu}, and sensitive to distributional shifts \cite{grigorescu2020survey, drenkow2021systematic}.  Additionally, the onboard computational resources for an ADAS or ADS are limited and shared among multiple safety systems. Therefore, in order to use DNNs in such a safety-critical resource-constrained environment, it is essential to efficiently produce uncertainty estimates. 

Uncertainty in machine learning (ML) arises from multiple sources, such as conflicting observations \cite{darling2019using}, variations in input data and sensor quality \cite{klas2018uncertainty}, and domain shifts \cite{farahani2021brief}. To enable DNNs to quantify uncertainty, various probabilistic deep learning (PDL) methods have been proposed. For example, these include methods that approximate the posterior distribution of a DNN's parameters given a training dataset (\eg, \cite{blundell2015weight, maddox2019simple, lakshminarayanan2017simple, gal2016dropout, wilson2020bayesian, izmailov2021bayesian, chen2014stochastic,neal2012bayesian,hoffman2013stochastic}). PDL methods that rely on sequentially sampling the DNN parameters to approximate the posterior distribution must produce a prediction for each sample. This increases computational effort, which can be prohibitive in a computationally limited environment, where the available computational resources are shared among multiple safety-critical functionalities. Similar to Monte Carlo sampling-based PDL methods, the computational cost of using ensembles of independently trained DNNs or Markov-Chain Monte Carlo (MCMC) samples also scales linearly with the number of predictions required to produce uncertainty estimates. 


PDL methods are typically evaluated on driving automation tasks, such as semantic segmentation, depth estimation and object detection tasks \cite{wang2025uncertainty, franchi2024make}, but less frequently for driver behavior or driving maneuver-related tasks, such as driver action or intention recognition \cite{vellenga2025llhmc}. Scaling PDL methods to large-scale real-world video datasets and DNN architectures remains challenging due to memory or latency constraints \cite{mukhoti2023deep, vellenga2024pthmc}. Last layer PDL (LL--PDL) methods produce uncertainty estimates more efficiently by reducing the number of stochastic parameters to the final layer of a DNN \cite{snoek2015scalable,watson2021latent}. For larger DNNs, computations for the last layer represent a negligible share of the total effort, which lowers the computational overhead of sequential sampling-based or ensemble methods for uncertainty estimation. However, while LL-PDL methods emphasize computational efficiency in uncertainty estimation, they generally overlook the potential benefits of incorporating the reconstruction error of the latent representation \cite{durasov2024zigzag,upadhyay2023probvlm, hornauer2023out}. 


To meet the demands of real-time decision-making in safety-critical and resource-constrained environments, driving automation components, such as video-based driver action and intention recognition, must produce reliable real-time predictions. In addition, under upcoming AI regulations (\eg, \cite{USA-AI, korea-AI, japan-AI, canada-AI, e2021proposal}), it is important to minimize undesirable behavior of AI systems in safety-critical environments. Therefore, in this paper we compare latent uncertainty representation (LUR) approaches to eight PDL approaches on in-distribution classification performance and uncertainty-based out-of-distribution (OOD) detection. We use four open-source video datasets for driver action or intention recognition (AssIstive Driving pErception [AIDE] dataset \cite{yang2023aide}, Brain4Cars [B4C] \cite{jain2015car}, ROad event Awareness Dataset \cite{singh2022road} [ROAD], and NuScenes \cite{caesar2020nuscenes}). Specifically for the NuScenes dataset, we provide and release new annotations: over 28,000 frame-level ego-vehicle driver action maneuver labels and 1,194 offline video driver intention labels.


\section{Related work}

\subsection{Driver action and intention recognition}
Action recognition concerns identifying what an agent, a driver in our case, currently does. Intention recognition refers to the identification of (immediate) future aspirations of an observed agent \cite{pereira2013state, sadri2011logic, vellenga2022driver}. An intention can be interpreted as a future plan, the purpose behind someone's action, or an intentional action \cite{setiya2009intention}. Thompson \cite{thompson2008naive} contends that intending to do something is not a fixed mental state, as it is neither permanent nor unchanging. It represents an ongoing process or state of incompleteness as we work toward fulfilling the intention. While a driver can possess multiple unpursued intentions simultaneously, we can only verify that someone might have intended to perform an action if it is eventually executed and observed.

Driver action recognition (DAR) is commonly performed online (\eg, \cite{cao2023e2e,chen2022gatehub,xu2019temporal,ramanishka2018toward,xu2021long,wang2021oadtr}). However, there are also examples of offline video-based DAR. Yang \etal (2023) \cite{yang2023aide} performed multiple driver and traffic scene related recognition tasks, including emotion recognition, facial landmark estimation, traffic context, behavior recognition and DAR, by fusing four different video streams. For video-based driver intention recognition (DIR), several offline approaches have been explored. For example, Gebert \etal (2019) \cite{gebert2019end} and Rong \etal (2020) \cite{rong2020driver} used optical flow features from in-cabin or external camera footage to improve DIR performance. Ma \etal (2023) \cite{ma2023cemformer} proposed a unified memory representation to jointly model in-cabin and external videos for DIR, and Vellenga \etal (2024) \cite{vellenga2024evaluation} fine-tuned a self-supervised Kinetics-400 pre-trained VideoMAE \cite{tong2022videomae} model on multiple DAR and DIR datasets. 

\subsection{Probabilistic deep learning}
Reliance on the softmax distribution is often inadequate for detecting OOD instances, as DNNs can exhibit high confidence even when uncertain \cite{gal2016dropout,guo2017calibration,van2020uncertainty}. To overcome this, numerous PDL methods have been proposed to estimate the predictive uncertainty  (\eg, \cite{zhou2021survey,jospin2022hands,arbel2023primer,gawlikowski2021survey,papamarkou2024position,manchingal2025position}). Currently, prevalent PDL approaches are MCMC, deep ensembles (DE), and variational inference \cite{papamarkou2024position}. MCMC approximates the posterior distribution of the DNN's parameters through sampling instead of optimization-based learning. MCMC is computationally expensive to use for both training and inference \cite{wenzel2020good, izmailov2021bayesian}. To balance the compute but still leverage some of the benefits of MCMC sampling, it is possible to restrict the sampling procedure to the last layer of the network \cite{vellenga2025llhmc}. Deep ensembles \cite{lakshminarayanan2017simple}, consisting of several randomly initialized independently trained DNNs, are easy to implement and can result in diverse predictions \cite{ashukhapitfalls}. However, both MCMC and ensemble-based approaches scale linearly with the number of predictions during training and inference, making them challenging to integrate into computationally limited environments. Using multiple classification layers (a sub-ensemble \cite{valdenegro2023sub}) can reduce computational requirements compared to full ensembles. To explicitly enforce diversity between multiple classification layers, Steger \etal \cite{steger2024function} proposed repulsive last layer ensembles (RLLE). 

Variational inference algorithms replace the DNN parameters with a set of variational distributions (\eg, 1-dimensional Gaussian distributions) to represent the DNN parameters. To approximate the true (intractable) posterior distribution, the evidence lower bound (ELBO) \cite{blei2017variational,jordan1999introduction} is used to minimize the Kullback-Leibler (KL) divergence between the variational distribution and the true posterior distribution. To enable backpropagation through a distribution, Bayes-by-backprop \cite{blundell2015weight} uses the reparameterization trick \cite{kingma2013auto}. At inference time, such stochastic variational inference (SVI) methods sample the parameters of the DNN from the approximated distributions. Such a procedure requires multiple forward passes to estimate the predictive uncertainty. The need to perform multiple forward passes can be undesirable for real-time predictions. However, multiple forward passes can lead to a more robust prediction, which is desirable in a safety-critical environment.

In resource-constrained environments, using ensembles or performing multiple forward passes may be undesirable. To avoid multiple forward passes, feature space distances or densities can be used as a sampling-free PDL approach \cite{van2020uncertainty,liu2020simple}.  For example, Deep Deterministic Uncertainty (DDU) \cite{mukhoti2023deep} demonstrated that performing class-wise Gaussian Discriminant Analysis on the latent representations of a pre-trained DNN with a regularized feature space can successfully identify OOD instances. Packed Ensemble (PE) \cite{laurent2023packed} parallelizes an ensemble into a single deterministic network. Grouped convolutions enable training multiple sub-networks simultaneously, which produces a diverse set of predictions. Variational Bayes Last Layer (VBBL) \cite{harrison2024variational} replaces the last layer of a DNN and leverages deterministic variational lower bound derivations on the marginal likelihood during training. This eliminates the need for sequential Monte Carlo sampling to estimate predictive uncertainty. 


\begin{figure*}[t]
    \centering
\includegraphics[trim={0cm 11.9cm 17.4cm 0.4cm},clip, width=0.75\linewidth, keepaspectratio]{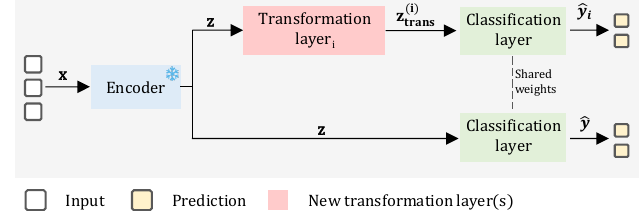}
        \caption{Schematic overview of the deterministic single inference latent uncertainty  representation. The raw input data $x$ is encoded into a latent representation $z$. This representation is then processed through the transformation layer(s) to produce the additional  latent representation(s) $z^{(i)}_{trans}$. Afterward, both \(z\) and  $z^{(i)}_{trans}$ are processed through the same classification layer $\theta_{LL}$, which produces $\hat{y}$ and $\hat{y_i}$. }
        \label{fig:LUR_schema}
    \end{figure*}


\section{Latent Uncertainty Representation}
As an alternative to ensemble, MCMC, or distance-based PDL methods, several studies used the reconstruction error to perform OOD detection \cite{zhou2022rethinking,hornauer2023out, durasov2024zigzag}. For example, Durasov \etal (2024) \cite{durasov2024zigzag} modify the input layer of the model and requires two predictions to estimate uncertainty. The first prediction is made without any prior information. This prediction is then appended to the input data and used for the second prediction. This approach assumes that in-distribution instances yield similar predictions ($\hat{y} \approx \hat{y}_i$). For OOD instances, this should result in outputs where $\hat{y}$ and $\hat{y}_i$ are dissimilar. However, Durasov's \etal (2024) \cite{durasov2024zigzag} approach has some undesirable properties. It requires modifying the input layer, and making two predictions, which can be impractical in safety-critical environments with limited computational resources.

Therefore, inspired by Durasov \etal (2024) \cite{durasov2024zigzag}, and as an alternative to LL--PDL methods, we propose using representation transformation layers. The goal of these layers is to produce additional latent representations that yield similar predictions for in-distribution instances (Figure \ref{fig:LUR_schema}). This approach follows the same reconstruction logic without changing the underlying (pre-trained) architecture. The additional computations required to produce extra latent representations are relatively negligible compared to producing the original latent representation $z$.

\subsection{Training the transformation layer(s)}
The most straightforward latent uncertainty representation (LUR) approach is to use a number of fully connected linear layers to transform the latent representation $z$ into multiple new latent representations. To train these randomly initialized linear transformation layers, the loss function is extended with an additional prediction loss for each latent representation $z^{(i)}_{trans}$ (Equation \ref{eq:pred_loss}) \cite{durasov2024zigzag}. In practice this means that every newly mapped latent representation $z^{(i)}_{trans}$ is also processed through the prediction layer $\theta_{LL}$.  The introduction of the transformation layers is conceptually similar to a sub-ensemble \cite{valdenegro2023sub}, but LUR creates multiple latent representations instead of relying on multiple classification layers. Similar to a sub-ensemble, this approach does not explicitly force diversity among the transformation layers.

\begin{equation}
    \mathcal{L}_{task} = \mathcal{L}(\theta_{LL}(z), y) +  \sum_{i=1}^n  \mathcal{L}(\theta_{LL}(z^{(i)}_{trans}), y),    
    \label{eq:pred_loss}
\end{equation}

where:
\begin{conditions} 
\mathcal{L}(\theta_{LL}(z), y) &=&  prediction loss based on the original\\
&&  latent representation $z$, \\
\mathcal{L}(\theta_{LL}(z^{(i)}_{trans}), y) &=&  prediction loss based on $z^{(i)}_{trans}$, \\
y&=& ground truth label, and\\
n&=&number of transformation layers. \\
\end{conditions}

Diverse DNN configurations might produce similar outcomes (e.g., the Rashomon effect \cite{breiman2001statistical}). This suggests that parameter diversity among transformation layers alone might not be sufficient to approximate the posterior. D’Angelo and Fortuin \cite{d2021repulsive} and Steger \etal \cite{steger2024function} demonstrated that adding a repulsive term in function space explicitly enforces diversity and can thereby improve uncertainty representations for OOD instances.

To explicitly enforce diversity among our transformation layers, we utilize a repulsive LUR (RLUR) training approach with particle-based optimization (POVI) \cite{d2021repulsive, steger2024function}. In contrast to SVI, which uses a simple parametric distribution $q(\theta)$ to approximate the posterior, POVI treats the particles (transformation layers) as a non-parametric distribution of the posterior. POVI uses an attraction term and repulsive term to update the parameters of the particles (Equation \ref{eq:rep_update}). The attraction term encourages the particles to move towards high-density regions of the posterior (toward better solutions). The repulsive term aims to prevent the transformation layers from collapsing into a single mode (the most likely region of the posterior distribution) by forcing the particles to explore different plausible posterior regions.


The choice of kernel function is important for the repulsive term, as it impacts how the particles (transformation layers) explore the posterior distribution. Following D’Angelo \& Fortuin \cite{d2021repulsive}, we use the Spectral Stein Gradient Estimator (SSGE) \cite{shi2018spectral} kernel, which considers the global structure of all particles and captures broader patterns in their relationships.


\begin{equation}
v(\theta_i) = \underbrace{\nabla_{\theta_i} \log p(\theta_i | \mathcal{D})}_{\text{Attraction}}
- \underbrace{\frac{\sum_{j=1}^{n} \nabla_{\theta_j} k (\theta_i, \theta_j)}{\sum_{j=1}^{n} k (\theta_i, \theta_j)}}_{\text{Repulsion}},
\label{eq:rep_update}
\end{equation}
where:
\begin{conditions}
p(\theta_i \mid \mathcal{D}) &=& posterior distribution of the\\
 && parameters $\theta_i$ given data $\mathcal{D}$, \\
\nabla_{\theta_i} \log p(\theta_i \mid \mathcal{D}) &=& gradient of the log-posterior with  \\
&& respect to the particle $\theta_i$, \\ 
     k(\cdot, \cdot)&=& kernel function to measure the  \\ 
     && similarity between the particles.\\
\end{conditions}

\section{Experimental setup}

\subsection{Problem formulation}
Suppose a driver is currently performing, or intends to perform, a driving maneuver $M \in \{m_1, \ldots, m_n\}$. For each maneuver $m_n$, a set of sensor observations $X = \{x_{1,1}, \ldots, x_{s,t}\} $ is collected from $s$ sensors for  $t$ time steps. The learning task for our model is to recognize the current or intended driving maneuver \( M \) based on the observed sequence $X$.

\subsection{Driver action and intention datasets}

\paragraph{AssIstive Driving pErception dataset (AIDE).}
The AIDE dataset \cite{yang2023aide} was collected in China and consists of four video streams (one in-cabin overview and three exterior views). The dataset consists of 2,898 driving scenarios, each lasting three seconds at 15 FPS, and comes with a train-validation-test split. We only use the driver action recognition labels (vehicle condition labels in \cite{yang2023aide}). The following five driving action labels are provided: \textit{ backward moving, forward moving, lane changing, parking,} and \textit{turning}.

\paragraph{Brain4Cars (B4C). }
{The Brain4Cars dataset \cite{jain2015car} is a DIR dataset collected in the United States. The dataset consists of 124 \textit{left lane changes}, 58 \textit{left turns}, 123 \textit{right lane changes}, 55 \textit{right turns}, and 234 \textit{driving straight maneuvers} with a five-fold 80/20 train/test split. The in-cabin camera operated at 30 FPS, while  the outside-facing camera operated at 25 FPS. }

\paragraph{ROad event Awareness Dataset (ROAD)}
The ROAD dataset \cite{singh2022road} is based on a subset of the Oxford RobotCar Dataset \cite{RobotCarDatasetIJRR}, and includes ego-vehicle and road user driving action labels. The original data collection vehicle was equipped with six cameras, LiDAR, and GPS sensors. Data collection took place between May 2014 and December 2015 in various weather conditions in the United Kingdom. The autonomous vehicle action labels from the ROAD dataset (\textit{AV-Stop, AV-Mov, AV-TurRht, AV-TurLft, AV-MovRht, AV-MovLft, AV-Ovtak}) are considered ego-vehicle driver action labels, as the vehicle was manually driven during data collection \cite{RobotCarDatasetIJRR}.

\paragraph{NuScenes. }
The NuScenes dataset \cite{caesar2020nuscenes} consists of driving scenes collected in Boston and Singapore using multiple exterior-facing cameras, LiDAR, RADAR, and GPS. The dataset provides a diverse set of contextual traffic scenarios and environments, as data was collected under varying weather conditions in both cities. Due to the geographical locations, the dataset includes scenarios where vehicles drive on both the left and right sides of the road. We annotated a subset of 28,549 frames with driving maneuver action labels (refer to Appendix \ref{app:annotation} for a detailed overview of our annotation protocol).

Driving in urban environments is frequently interrupted by waiting for traffic lights or other road users during maneuvers. Similar to Singh \etal \cite{singh2022road}, we include \textit{waiting} as one of the driving maneuvers. To avoid ambiguous or extremely rare maneuvers, we exclude actions such as \textit{u-turns}, \textit{move-left}, or \textit{move-right} \cite{ramanishka2018toward,singh2022road}. We only include driving scenes in which the intended maneuver was observed and at least eight frames in the preceding ten seconds are available. As a result, our offline NuScenes driving intention maneuver benchmark consists of 88 \textit{right lane changes}, 109 \textit{left lane changes}, 133 \textit{right turns}, 127 \textit{left turns}, 220 \textit{waiting}, and 517 \textit{forward moving} scenarios. Using the original NuScenes split resulted in a 67/33 train/test split.


 \subsection{Metrics}

\subsubsection{Classification performance and calibration}
We report the accuracy $(\uparrow)$ and F1-score $(\uparrow)$, consistent with previous offline DAR and DIR studies. To assess the performance-confidence calibration of the models, we report the ten bins adaptive calibration error (ACE, $\downarrow$) \cite{nixon2019measuring}. 

\subsubsection{Uncertainty quantification and calibration}
While decomposing uncertainty into aleatoric (inherent randomness) and epistemic (lack of knowledge) components \cite{der2009aleatory} can help identify model blind spots, the actual disentanglement remains challenging in practice \cite{mucsanyi2024benchmarking}. Therefore, we rely on the predictive entropy \cite{smith2018understanding}, which captures both types of uncertainty. Intuitively, for an instance $x$, the predictive entropy reaches its maximum value when the predicted probabilities for all classes are equal. The higher the predicted probability for a single class, the lower the predictive entropy, indicating greater certainty about the produced predictions.

We use the Relative Area Under the Lift Curve (rAULC, $\uparrow$) \cite{postels2022practicality} to measure the effectiveness of a model's uncertainty estimates in distinguishing between correctly and incorrectly classified in-distribution instances. Intuitively, a higher rAULC indicates that the model can better distinguish between correct and incorrect predictions based on its uncertainty estimates.

 

\subsubsection{Out-of-distribution detection}

A PDL method should be able to distinguish between in-distribution and OOD samples. To evaluate whether OOD instances yield high uncertainty estimates, we compute the areas under the Receiver Operating Characteristic (ROC, $\uparrow$) and Precision-Recall (PR,$\uparrow$) curves \cite{malinin2018predictive,durasov2024zigzag}. Lastly, the False Positive Rate at 95\% Recall (FPR95, $\downarrow$) is widely used to assess the performance of OOD detection methods (\eg, \cite{laurent2023packed,franchi2024make}). FPR95 indicates the rate at which OOD samples are incorrectly classified as in-distribution when the model correctly identifies 95\% of in-distribution samples. 


\subsection{Underlying model architecture}
For the AIDE, B4C, ROAD, and NuScenes datasets, we use the implementation from \cite{vellenga2025llhmc}. We use a self-supervised Kinetics-400 pre-trained Vision-Transformer (ViT)-Base architecture that expects 16 frame videos with a resolution of 224 by 224 pixels, has 12 hidden layers with a dimension of 768, and 12 attention heads \cite{arnab2021vivit,tong2022videomae}. We freeze the input projection layer and fine-tune only the attention layers of the pre-trained ViT, as recommended by Touvron \etal \cite{touvron2022three}. For the AIDE and B4C datasets, which include multiple synchronized modalities, we fuse latent representations via attentional feature fusion \citep{dai2021attentional}. 


\subsection{Implementation details}
To train the underlying model responsible for producing latent representations, we randomly sample 16 frames from each video clip or apply zero padding to shorter clips during training. At test time, we use evenly spaced sampling of 16 video frames based on the video length. We use the AdamW optimizer \cite{loshchilov2018decoupled} with a weight decay factor of 0.05, and train the models for 20 epochs with early stopping (patience of five epochs). Except for the deep ensemble, each PDL method relies on the same latent representations. For the LL--PDL methods, we perform grid search over different random seeds to evaluate the number of predictions (5, 10, 15, 20, 25, 30, 35, 40, 45, 50), batch size (16, 32, 64), and learning rate ($1\mathrm{e}{-2}, 1\mathrm{e}{-3}, 1\mathrm{e}{-4}$) for different numbers of epochs (5, 10, 15, 20, 25). For the LL--HMC hyperparameters (burn-in samples, prior standard deviations, target acceptance, number of chains and number of collected samples) we follow the ranges as \cite{vellenga2025llhmc}. 

\begin{table*}[t]
\centering
\caption{Average in-distribution, video-based driver action and intention recognition performance on the AIDE, B4C, ROAD, and NuScenes datasets. Results are reported for the top-performing hyperparameter configurations, averaged over five random seeds, with two standard errors of the mean. The LL--PDL and LUR methods use the latent representations produced by the \textit{Regular} model. Metrics include Acc (Accuracy), ACE (Adaptive Calibration Error), and rAULC (relative Area Under the Lifted Curve). The best results are highlighted in \colorbox{best}{blue}, and the second best in \colorbox{2best}{light blue}.}
\vspace{-0.10cm}
\resizebox{0.98\linewidth}{!}{
\begin{tabular}{l|c|c|c|c||c|c|c|c||c|c|c|c||c|c|c|c}
\toprule
 & \multicolumn{4}{c||}{\textbf{AIDE} \cite{yang2023aide}} & \multicolumn{4}{c||}{\textbf{B4C} \cite{jain2015car}} & \multicolumn{4}{c||}{\textbf{ROAD} \cite{singh2022road}}  & \multicolumn{4}{c}{\textbf{NuScenes} \cite{caesar2020nuscenes}}  \\
 & \textbf{Acc ($\uparrow$)} & \textbf{F1 ($\uparrow$)} & \textbf{ACE ($\downarrow$)} & \textbf{rAULC ($\uparrow$)} & \textbf{Acc ($\uparrow$)} &  \multicolumn{1}{|c|}{\textbf{F1 ($\uparrow$)}} & \textbf{ACE ($\downarrow$)} & \textbf{rAULC ($\uparrow$)} & \textbf{Acc ($\uparrow$)} & \textbf{F1 ($\uparrow$)} & \textbf{ACE ($\downarrow$)} & \textbf{rAULC ($\uparrow$)} & \textbf{Acc ($\uparrow$)} & \textbf{F1 ($\uparrow$)} & \textbf{ACE ($\downarrow$)} & \textbf{rAULC ($\uparrow$)} \\
\midrule
 \textbf{Regular}  & 87.19 & 81.88 & \cellcolor{2best} 0.027 & 0.77 & 92.31 & 93.71 & 0.050 & 0.53 & 71.64 & 45.36 & \cellcolor{best} 0.063 &\cellcolor{2best} 0.61 & 58.81 & 44.63 & 0.077 & 0.56 \\
  \textbf{DE (N=5)}  &\cellcolor{best} 89.66 & 82.24 &  \cellcolor{best} 0.019 & \cellcolor{2best} 0.80 & 90.88 & 91.49 & 0.059 & \cellcolor{best} 0.73 & \cellcolor{best} 74.24 & 44.57 &  0.066 & \cellcolor{best} 0.70 & 60.36 & 43.46&  \cellcolor{best} 0.062 & 0.53 \\
  \textbf{DDU} & 57.31 & 19.12 & 0.088 &\cellcolor{best}  0.85 & 76.07 & 74.72& 0.124 & 0.61 & 59.70 & 23.85 & 0.138 & 0.42 & 52.07 & 21.24 & 0.100 & 0.57  \\ \midrule
\textbf{BBB--LL} & 88.67$\pm$0.36 &  83.13$\pm$0.26 &  0.032$\pm$0.005 & 0.73$\pm$0.02&    93.16$\pm$0.94 &94.28$\pm$0.87 &0.025$\pm$0.007 & 0.65$\pm$0.08  & 71.64$\pm$3.13 & \cellcolor{2best}56.14$\pm$4.37 & \cellcolor{2best} 0.068$\pm$0.003 &  0.50$\pm$0.13 & 58.81$\pm$0.94 & 44.70$\pm$1.41 & 0.120$\pm$0.001 & 0.52$\pm$0.03\\

\textbf{PE--LL} & 88.64$\pm$0.32 &\cellcolor{2best}83.37$\pm$0.02 &  0.030$\pm$0.003 & 0.74$\pm$0.01 &\cellcolor{2best} 94.02$\pm$2.24  &   \cellcolor{2best} 95.02$\pm$0.02 & \cellcolor{best}  0.022$\pm$0.005 & 0.67$\pm$0.03 & 68.66$\pm$2.11 & \cellcolor{best} 59.28$\pm$1.38 & 0.077$\pm$0.008 &  0.36$\pm$0.09 & \cellcolor{2best} 61.09$\pm$0.19 & \cellcolor{2best} 45.74$\pm$0.15 & 0.090$\pm$0.001 & 0.56$\pm$0.01 \\

\textbf{SE} & 88.31$\pm$0.19 &  82.69$\pm$0.21 &  0.029$\pm$0.000 & 0.75$\pm$0.01 & 93.16$\pm$0.18 &  94.38$\pm$0.04 & \cellcolor{2best}  0.023$\pm$0.001 & \cellcolor{2best}  0.72$\pm$0.00 &  68.06$\pm$2.23 & 45.63$\pm$2.45 & \cellcolor{2best} 0.065$\pm$0.002 &  0.57$\pm$0.01 & 60.62$\pm$0.28 & 45.42$\pm$0.26 & 0.080$\pm$0.000& 0.55$\pm$0.01 \\

\textbf{VBLL} & 88.67$\pm$0.36 &  83.13$\pm$0.26 &  0.032$\pm$0.005 &  0.73$\pm$0.02 &  92.14$\pm$1.26 &  93.47$\pm$1.13 & 0.116$\pm$0.038 &  0.55$\pm$0.10 & \cellcolor{2best}72.54$\pm$2.60 &  47.25$\pm$2.08 & 0.084$\pm$0.004 &  0.48$\pm$0.05 & 61.04$\pm$0.88 & 44.91$\pm$0.66 & 0.059$\pm$0.001& \cellcolor{2best} 0.58$\pm$0.02\\

\textbf{LL--HMC} & \cellcolor{2best} 89.00$\pm$0.16 & \cellcolor{best} 83.87$\pm$0.08 & 0.029$\pm$0.002 & 0.68$\pm$0.06 & \cellcolor{2best} 94.02$\pm$0.42 &  95.00$\pm$0.29 & 0.027$\pm$0.007 & 0.66$\pm$0.10 &   70.15$\pm$5.05 &   54.18$\pm$4.43 & 0.073$\pm$0.014 & 0.41$\pm$0.12 & \cellcolor{best} 61.45$\pm$0.95 & \cellcolor{best} 46.55$\pm$0.48 & \cellcolor{2best} 0.068$\pm$0.002 & 0.57$\pm$0.02  \\

\textbf{RLLE} & 88.80$\pm$0.22 & 81.43$\pm$0.68 & 0.031$\pm$0.001 & 0.56$\pm$0.06	&	88.21$\pm$1.47 & 88.68$\pm$2.26 & 0.024$\pm$0.010 & 0.41$\pm$0.09	&	64.78$\pm$4.49 & 33.64$\pm$5.03 & 0.170$\pm$0.012 & 0.50$\pm$0.14	&	60.57$\pm$0.78 & 38.52$\pm$1.78 & 0.124$\pm$0.004 & \cellcolor{best} 0.60$\pm$0.02	\\

\textbf{LUR} & 	88.87$\pm$0.32 & 83.24$\pm$0.18 & 0.060$\pm$0.004 & 0.72$\pm$0.03	&	\cellcolor{best} 94.19$\pm$0.34 & \cellcolor{best} 95.16$\pm$0.29 & 0.032$\pm$0.001 & 0.72$\pm$0.06	&	68.66$\pm$1.33 & 55.82$\pm$2.01 & 0.081$\pm$0.001 & 0.46$\pm$0.12	&	60.52$\pm$0.13 & 45.40$\pm$0.07 & 0.080$\pm$0.001 & 0.56$\pm$0.00 \\
\textbf{RLUR} & 88.44$\pm$0.22 & 82.58$\pm$0.27 & 0.030$\pm$0.000& 0.72$\pm$0.02	&	93.68$\pm$0.42 & 94.75$\pm$0.32 & 0.030$\pm$0.001& 0.70$\pm$0.04	&	65.97$\pm$1.74 & 42.40$\pm$2.57 & 0.080$\pm$0.010 & 0.51$\pm$0.08	&	59.12$\pm$0.45 & 43.68$\pm$0.65 & 0.120$\pm$0.001 & 0.51$\pm$0.02  \\

 \bottomrule
\end{tabular}
}
\label{tab:id_perf}
\end{table*}

\begin{table*}[!ht]
\centering
\caption{Average uncertainty-based OOD detection results and two standard errors of the mean for the AIDE, B4C, ROAD, and NuScenes datasets. Results are averaged across five random seeds using the top-performing hyperparameter configurations. ROC-AUC=Receiver Operating Characteristic - Area Under the Curve, PR=Precision-Recall - Area Under the Curve, FPR95=False Positive Rate at 95\% True Positive Rate. } 
\vspace{-0.1cm}
\resizebox{\textwidth}{!}{
\begin{tabular}{l|c|c|c|c|c|c||c|c|c|c|c|c||c|c|c||c|c|c|c|c|c}
\toprule
\multicolumn{1}{c||}{} & \multicolumn{6}{c||}{\textbf{AIDE} \cite{yang2023aide}} & \multicolumn{6}{c||}{\textbf{B4C} \cite{jain2015car}} & \multicolumn{3}{c||}{\textbf{ROAD} \cite{singh2022road}} & \multicolumn{6}{c}{\textbf{NuScenes} \cite{caesar2020nuscenes}} \\
\cmidrule(lr){2-7}\cmidrule(lr){8-13}\cmidrule(lr){14-16}\cmidrule(lr){17-22}
\multicolumn{1}{c||}{} & \multicolumn{3}{c|}{\textbf{OOD min}} & \multicolumn{3}{c||}{\textbf{OOD max}} & \multicolumn{3}{c|}{\textbf{OOD min}} & \multicolumn{3}{c||}{\textbf{OOD max}} & \multicolumn{3}{c||}{\textbf{OOD min}} & \multicolumn{3}{c|}{\textbf{OOD min}} & \multicolumn{3}{c}{\textbf{OOD max}} \\
\multicolumn{1}{c||}{\textbf{}} & 
\begin{tabular}{@{}c@{}}\textbf{ROC-}\\ \textbf{AUC} ($\uparrow$)\end{tabular} & 
\begin{tabular}{@{}c@{}}\textbf{PR-}\\\textbf{AUC} ($\uparrow$)\end{tabular} & 
\begin{tabular}{@{}c@{}}\textbf{FPR95}\\($\downarrow$)\end{tabular} & 
\begin{tabular}{@{}c@{}}\textbf{ROC-}\\\textbf{AUC} ($\uparrow$)\end{tabular} & 
\begin{tabular}{@{}c@{}}\textbf{PR-}\\\textbf{AUC} ($\uparrow$)\end{tabular} & 
\begin{tabular}{@{}c@{}}\textbf{FPR95}\\($\downarrow$)\end{tabular} & 
\begin{tabular}{@{}c@{}}\textbf{ROC-}\\\textbf{AUC} ($\uparrow$)\end{tabular} & 
\begin{tabular}{@{}c@{}}\textbf{PR-}\\\textbf{AUC} ($\uparrow$)\end{tabular} & 
\begin{tabular}{@{}c@{}}\textbf{FPR95}\\($\downarrow$)\end{tabular} & 
\begin{tabular}{@{}c@{}}\textbf{ROC-}\\\textbf{AUC} ($\uparrow$)\end{tabular} & 
\begin{tabular}{@{}c@{}}\textbf{PR-}\\\textbf{AUC} ($\uparrow$)\end{tabular} & 
\begin{tabular}{@{}c@{}}\textbf{FPR95}\\($\downarrow$)\end{tabular} & 
\begin{tabular}{@{}c@{}}\textbf{ROC-}\\\textbf{AUC} ($\uparrow$)\end{tabular} & 
\begin{tabular}{@{}c@{}}\textbf{PR-}\\\textbf{AUC} ($\uparrow$)\end{tabular} & 
\begin{tabular}{@{}c@{}}\textbf{FPR95}\\($\downarrow$)\end{tabular} & 
\begin{tabular}{@{}c@{}}\textbf{ROC-}\\\textbf{AUC} ($\uparrow$)\end{tabular} & 
\begin{tabular}{@{}c@{}}\textbf{PR-}\\\textbf{AUC} ($\uparrow$)\end{tabular} & 
\begin{tabular}{@{}c@{}}\textbf{FPR95}\\($\downarrow$)\end{tabular} & 
\begin{tabular}{@{}c@{}}\textbf{ROC-}\\\textbf{AUC} ($\uparrow$)\end{tabular} & 
\begin{tabular}{@{}c@{}}\textbf{PR-}\\\textbf{AUC} ($\uparrow$)\end{tabular} & 
\begin{tabular}{@{}c@{}}\textbf{FPR95}\\($\downarrow$)\end{tabular} \\

\midrule
\textbf{Regular} & 0.55 & 0.19 & 0.91 & 0.82 & 0.94 & 0.41 & 0.51 & 0.37 & 0.90 & 0.64 & 0.82 & 0.84 & 0.51 & 0.37 & 0.90 & 0.19 & 0.25 & 1.00 & 0.63 & 0.78 & 0.84 \\
\textbf{DE (N=5)} & 0.69 & 0.28 & 0.85 & 0.84 & 0.94 & 0.35 & 0.52 & 0.39 & 0.90 & 0.79 & 0.90 & 0.45 & 0.55 & 0.41 & 0.88 & 0.60 & 0.24 & 0.89 & 0.67 & 0.81 & 0.79 \\
\textbf{DDU} & 0.73 & 0.31 & 0.49 & 0.80 & 0.90 & 0.45 & 0.63 & 0.44 & 0.87 & 0.48 & 0.75 & 0.96 & 0.63 & 0.44 & 0.87 & 0.34 & 0.31 & 0.97 & 0.50 & 0.68 & 0.95 \\ \midrule
\textbf{BBB--LL} &  0.95$\pm$0.02 & 0.72$\pm$0.08 & 0.10$\pm$0.05	&  0.96$\pm$0.01 &  0.99$\pm$0.01 &  0.22$\pm$0.06	& 0.84$\pm$0.03 &  0.70$\pm$0.04 & 0.48$\pm$0.21	&  0.80$\pm$0.02 & 0.92$\pm$0.01 &  0.69$\pm$0.10 & 	0.60$\pm$0.10 & 0.49$\pm$0.13 & 0.85$\pm$0.06 & 0.76$\pm$0.01 & 0.53$\pm$0.03 & 0.78$\pm$0.04 & 0.78$\pm$0.02 & 0.89$\pm$0.01 & 0.72$\pm$0.04  \\

\textbf{PE--LL} &  0.55$\pm$0.01 & 0.52$\pm$0.01 & 0.83$\pm$0.01	& 0.56$\pm$0.01 & 0.53$\pm$0.01 & 0.72$\pm$0.02 &	0.52$\pm$0.01 & 0.52$\pm$0.01 & 0.91$\pm$0.01 &	0.53$\pm$0.01 & 0.53$\pm$0.00 & 0.94$\pm$0.02 &	0.53$\pm$0.01 &  0.52$\pm$0.01 & 0.92$\pm$0.01 & 0.52$\pm$0.01 & 0.52$\pm$0.00 & 0.93$\pm$0.00 & 0.52$\pm$0.00& 0.52$\pm$0.00& 0.93$\pm$0.01 \\
\textbf{SE} &  0.59$\pm$0.01 & 0.21$\pm$0.01 & 0.89$\pm$0.01 & 	0.88$\pm$0.01 & 0.96$\pm$0.01 & 0.41$\pm$0.01	& 0.72$\pm$0.02 & 0.54$\pm$0.01 & 0.50$\pm$0.02 &	0.71$\pm$0.01 & 0.88$\pm$0.01 & 0.79$\pm$0.01 &	0.45$\pm$0.05 & 0.33$\pm$0.02 & 0.86$\pm$0.10 & 0.6$\pm$0.01 & 0.26$\pm$0.01 & 0.91$\pm$0.01 & 0.69$\pm$0.01 & 0.82$\pm$0.00& 0.77$\pm$0.03  \\

\textbf{VBLL} & 0.66$\pm$0.04 & 0.25$\pm$0.03 & 0.79$\pm$0.08	&0.89$\pm$0.01 & 0.97$\pm$0.01 & 0.42$\pm$0.04	& 0.82$\pm$0.05 & 0.64$\pm$0.07 &  0.41$\pm$0.09 &	0.65$\pm$0.02 & 0.84$\pm$0.02 & 0.88$\pm$0.06 &	0.57$\pm$0.09 & 0.44$\pm$0.11 & 0.80$\pm$0.05 & 0.60$\pm$0.02 & 0.26$\pm$0.01 & 0.95$\pm$0.04 & 0.69$\pm$0.01 & 0.81$\pm$0.00 & 0.76$\pm$0.03  \\
 
\textbf{LL--HMC} & 0.91$\pm$0.03 &  0.58$\pm$0.10 & 0.18$\pm$0.05 & 0.95$\pm$0.01 & \cellcolor{2best} 0.99$\pm$0.01 &   0.14$\pm$0.01	&    0.85$\pm$0.04 &    0.73$\pm$0.05 &  0.36$\pm$0.08&  0.91$\pm$0.01 &    \cellcolor{2best} 0.97$\pm$0.01 &    0.39$\pm$0.05& \cellcolor{2best}  0.87$\pm$0.01 &  \cellcolor{2best} 0.72$\pm$0.01 &   0.31$\pm$0.04 & 0.86$\pm$0.01 & 0.68$\pm$0.01 & 0.57$\pm$0.02  &0.94$\pm$0.00 & 0.97$\pm$0.01 & 0.26$\pm$0.02
 \\
 \textbf{RLLE} &\cellcolor{2best} 0.99$\pm$0.00& \cellcolor{2best} 0.96$\pm$0.02 & \cellcolor{2best} 0.02$\pm$0.00	& \cellcolor{best} 1.00$\pm$0.01 & \cellcolor{best}1.00$\pm$0.00&  \cellcolor{2best} 0.02$\pm$0.02	&	\cellcolor{best}0.99$\pm$0.01 & \cellcolor{best}0.99$\pm$0.02 &\cellcolor{best} 0.04$\pm$0.06	& \cellcolor{best} 1.00$\pm$0.01 &\cellcolor{best} 1.00$\pm$0.00& \cellcolor{best} 0.02$\pm$0.02	& \cellcolor{best}1.00$\pm$0.00& \cellcolor{best}1.00$\pm$0.01 & \cellcolor{best} 0.01$\pm$0.01	& \cellcolor{2best} 0.98$\pm$0.01 & \cellcolor{best} 0.94$\pm$0.03 & \cellcolor{2best} 0.09$\pm$0.05	& \cellcolor{best}1.00$\pm$0.00&\cellcolor{best}1.00$\pm$0.00&\cellcolor{best} 0.00$\pm$0.00\\
 
  \textbf{LUR} & \cellcolor{best} 1.00$\pm$0.00& \cellcolor{best} 0.97$\pm$0.01 & \cellcolor{best} 0.00$\pm$0.00	& \cellcolor{best} 1.00$\pm$0.00& \cellcolor{best} 1.00$\pm$0.00& \cellcolor{best} 0.01$\pm$0.01	&	\cellcolor{2best} 0.97$\pm$0.00&  \cellcolor{2best} 0.91$\pm$0.01 & \cellcolor{2best} 0.05$\pm$0.01	& \cellcolor{2best}	 0.99$\pm$0.00& \cellcolor{best} 1.00$\pm$0.00&\cellcolor{2best} 0.04$\pm$0.02	&\cellcolor{best} 1.00$\pm$0.00&\cellcolor{best}1.00$\pm$0.00&\cellcolor{best} 0.00$\pm$0.00	& \cellcolor{best}	0.99$\pm$0.00 & \cellcolor{2best} 0.93$\pm$0.01 & \cellcolor{best} 0.03$\pm$0.00	& \cellcolor{2best}	0.99$\pm$0.01 & \cellcolor{2best} 0.99$\pm$0.00& \cellcolor{2best} 0.04$\pm$0.01\\
 
  \textbf{RLUR} & 0.92$\pm$0.02 & 0.61$\pm$0.05 & 0.20$\pm$0.05	&	\cellcolor{2best}0.97$\pm$0.00& \cellcolor{2best} 0.99$\pm$0.00& 0.10$\pm$0.01	&	0.81$\pm$0.02 & 0.68$\pm$0.01 & 0.47$\pm$0.05	&	0.90$\pm$0.01 & 0.96$\pm$0.00& 0.35$\pm$0.05	&	0.84$\pm$0.03 & 0.69$\pm$0.05 & 0.4$\pm$0.05	&	0.85$\pm$0.00& 0.65$\pm$0.02 & 0.61$\pm$0.03	&	0.95$\pm$0.00& 0.98$\pm$0.00& 0.24$\pm$0.02 \\

\bottomrule
\end{tabular}
}
\label{tab:ood_perf}
\end{table*}

\begin{table*}[!ht]
\centering
\caption{Best average hyperparameter configuration OOD detection results and two standard errors of the mean grid search results across five random seeds. }

\vspace{-0.1cm}
\resizebox{\textwidth}{!}{
\begin{tabular}{l|c|c|c|c|c|c||c|c|c|c|c|c||c|c|c||c|c|c|c|c|c}
\toprule
\multicolumn{1}{c||}{} & \multicolumn{6}{c||}{\textbf{AIDE} \cite{yang2023aide}} & \multicolumn{6}{c||}{\textbf{B4C} \cite{jain2015car}} & \multicolumn{3}{c||}{\textbf{ROAD} \cite{singh2022road}} & \multicolumn{6}{c}{\textbf{NuScenes} \cite{caesar2020nuscenes}} \\
\cmidrule(lr){2-7}\cmidrule(lr){8-13}\cmidrule(lr){14-16}\cmidrule(lr){17-22}
\multicolumn{1}{c||}{} & \multicolumn{3}{c|}{\textbf{OOD min}} & \multicolumn{3}{c||}{\textbf{OOD max}} & \multicolumn{3}{c|}{\textbf{OOD min}} & \multicolumn{3}{c||}{\textbf{OOD max}} & \multicolumn{3}{c||}{\textbf{OOD min}} & \multicolumn{3}{c|}{\textbf{OOD min}} & \multicolumn{3}{c}{\textbf{OOD max}} \\
\multicolumn{1}{c||}{\textbf{}} & 
\begin{tabular}{@{}c@{}}\textbf{ROC-}\\ \textbf{AUC} ($\uparrow$)\end{tabular} & 
\begin{tabular}{@{}c@{}}\textbf{PR-}\\\textbf{AUC} ($\uparrow$)\end{tabular} & 
\begin{tabular}{@{}c@{}}\textbf{FPR95}\\($\downarrow$)\end{tabular} & 
\begin{tabular}{@{}c@{}}\textbf{ROC-}\\\textbf{AUC} ($\uparrow$)\end{tabular} & 
\begin{tabular}{@{}c@{}}\textbf{PR-}\\\textbf{AUC} ($\uparrow$)\end{tabular} & 
\begin{tabular}{@{}c@{}}\textbf{FPR95}\\($\downarrow$)\end{tabular} & 
\begin{tabular}{@{}c@{}}\textbf{ROC-}\\\textbf{AUC} ($\uparrow$)\end{tabular} & 
\begin{tabular}{@{}c@{}}\textbf{PR-}\\\textbf{AUC} ($\uparrow$)\end{tabular} & 
\begin{tabular}{@{}c@{}}\textbf{FPR95}\\($\downarrow$)\end{tabular} & 
\begin{tabular}{@{}c@{}}\textbf{ROC-}\\\textbf{AUC} ($\uparrow$)\end{tabular} & 
\begin{tabular}{@{}c@{}}\textbf{PR-}\\\textbf{AUC} ($\uparrow$)\end{tabular} & 
\begin{tabular}{@{}c@{}}\textbf{FPR95}\\($\downarrow$)\end{tabular} & 
\begin{tabular}{@{}c@{}}\textbf{ROC-}\\\textbf{AUC} ($\uparrow$)\end{tabular} & 
\begin{tabular}{@{}c@{}}\textbf{PR-}\\\textbf{AUC} ($\uparrow$)\end{tabular} & 
\begin{tabular}{@{}c@{}}\textbf{FPR95}\\($\downarrow$)\end{tabular} & 
\begin{tabular}{@{}c@{}}\textbf{ROC-}\\\textbf{AUC} ($\uparrow$)\end{tabular} & 
\begin{tabular}{@{}c@{}}\textbf{PR-}\\\textbf{AUC} ($\uparrow$)\end{tabular} & 
\begin{tabular}{@{}c@{}}\textbf{FPR95}\\($\downarrow$)\end{tabular} & 
\begin{tabular}{@{}c@{}}\textbf{ROC-}\\\textbf{AUC} ($\uparrow$)\end{tabular} & 
\begin{tabular}{@{}c@{}}\textbf{PR-}\\\textbf{AUC} ($\uparrow$)\end{tabular} & 
\begin{tabular}{@{}c@{}}\textbf{FPR95}\\($\downarrow$)\end{tabular} \\

\midrule
\textbf{BBB--LL} & 0.55$\pm$0.11 & 0.20$\pm$0.04 & 0.84$\pm$0.15	&	0.80$\pm$0.06 & 0.93$\pm$0.02 & 0.58$\pm$0.20	&	0.32$\pm$0.05 & 0.30$\pm$0.02 & 0.99$\pm$0.03	&	 0.50$\pm$0.11 & 0.76$\pm$0.07 & 0.92$\pm$0.04	&	0.29$\pm$0.06 & 0.28$\pm$0.02 & 0.97$\pm$0.04 & 0.59$\pm$0.04 & 0.24$\pm$0.03 & 0.93$\pm$0.04  & 0.59$\pm$0.06 & 0.76$\pm$0.04 & 0.88$\pm$0.04   \\

\textbf{PE--LL} &  0.50$\pm$0.00& \cellcolor{2best} 0.50$\pm$0.00& 0.95$\pm$0.01	&	0.51$\pm$0.00& 0.51$\pm$0.00& 0.93$\pm$0.01	&	0.51$\pm$0.00& 0.51$\pm$0.00& 0.95$\pm$0.01	&	0.52$\pm$0.01 & 0.51$\pm$0.01 & 0.94$\pm$0.01	&	0.51$\pm$0.00& \cellcolor{2best} 0.51$\pm$0.00& 0.95$\pm$0.00 & 0.51$\pm$0.00& 0.51$\pm$0.00& 0.95$\pm$0.00 & 0.51$\pm$0.00& 0.51$\pm$0.00& 0.95$\pm$0.00\\

\textbf{SE} &  0.51$\pm$0.01 & 0.17$\pm$0.00& 0.86$\pm$0.02	&	0.81$\pm$0.00& 0.93$\pm$0.00& 0.41$\pm$0.01	&	0.29$\pm$0.00& 0.29$\pm$0.00& 1.00$\pm$0.00	&	0.32$\pm$0.02 & 0.69$\pm$0.01 & 0.99$\pm$0.01	&	0.17$\pm$0.01 & 0.24$\pm$0.00& 1.00$\pm$0.00 & 0.53$\pm$0.01 & 0.20$\pm$0.01 & 0.93$\pm$0.01 & 0.65$\pm$0.01 & 0.79$\pm$0.00& 0.84$\pm$0.01 \\

\textbf{VBLL} &  0.44$\pm$0.02 & 0.14$\pm$0.01 & 0.96$\pm$0.02	&	0.54$\pm$0.14 & 0.84$\pm$0.07 & 0.88$\pm$0.13	&	0.49$\pm$0.10& 0.36$\pm$0.07 & 0.90$\pm$0.12	&	0.48$\pm$0.03 & 0.74$\pm$0.02 & 0.96$\pm$0.02	&	0.16$\pm$0.03 & 0.24$\pm$0.01 & 0.99$\pm$0.01 & 0.48$\pm$0.05 & 0.18$\pm$0.02 & 0.95$\pm$0.02 & 0.50$\pm$0.02 & 0.70$\pm$0.01 & 0.95$\pm$0.02\\

\textbf{LL--HMC} & 0.53$\pm$0.02 & 0.18$\pm$0.01 & 0.87$\pm$0.05	&	0.82$\pm$0.06 & 0.94$\pm$0.02 & 0.48$\pm$0.20	&	0.57$\pm$0.06 & 0.42$\pm$0.04 & 0.78$\pm$0.10	&	0.57$\pm$0.06 & 0.42$\pm$0.04 & 0.78$\pm$0.10	&0.28$\pm$0.08 & 0.27$\pm$0.02 & 0.97$\pm$0.03 & 0.59$\pm$0.01 & 0.23$\pm$0.01 & 0.93$\pm$0.01 &0.67$\pm$0.02 & 0.80$\pm$0.01 & 0.80$\pm$0.01 \\

\textbf{RLLE} & \cellcolor{best} 0.84$\pm$0.09 & \cellcolor{best}  0.57$\pm$0.13 &  \cellcolor{best} 0.42$\pm$0.14& \cellcolor{best} 0.94$\pm$0.03 & \cellcolor{best} 0.98$\pm$0.01 & \cellcolor{best} 0.20$\pm$0.06 & \cellcolor{best} 0.91$\pm$0.08 & \cellcolor{best} 0.85$\pm$0.10 &  \cellcolor{best} 0.25$\pm$0.19 & \cellcolor{best} 0.95$\pm$0.02 & \cellcolor{best} 0.98$\pm$0.01 & \cellcolor{best} 0.17$\pm$0.09 & \cellcolor{best} 0.92$\pm$0.10 & \cellcolor{best} 0.85$\pm$0.17 & \cellcolor{best} 0.17$\pm$0.17 & \cellcolor{best} 0.91$\pm$0.04 & \cellcolor{best} 0.79$\pm$0.08 &  \cellcolor{best}  0.44$\pm$0.21 &  \cellcolor{best} 0.92$\pm$0.10 &  \cellcolor{best}  0.85$\pm$0.17 &   \cellcolor{best} 0.17$\pm$0.17
 \\
\textbf{LUR} &  \cellcolor{2best} 0.83$\pm$0.01 & 0.24$\pm$0.04 & 0.72$\pm$0.03&0.81$\pm$0.01 & 0.93$\pm$0.00& 0.46$\pm$0.00& 0.50$\pm$0.01 & 0.36$\pm$0.01 & 0.73$\pm$0.03 & 0.61$\pm$0.00& 0.80$\pm$0.00& 0.92$\pm$0.01 & 0.16$\pm$0.00& 0.24$\pm$0.00& 1.00$\pm$0.00& 0.57$\pm$0.00& 0.22$\pm$0.00& 0.95$\pm$0.00& 0.67$\pm$0.01 & 0.79$\pm$0.01 & 0.64$\pm$0.01 \\

\textbf{RLUR} &0.81$\pm$0.03 & 0.41$\pm$0.05 & \cellcolor{2best} 0.44$\pm$0.01  & \cellcolor{2best} 0.91$\pm$0.01 &  \cellcolor{2best} 0.97$\pm$0.00& \cellcolor{2best} 0.27$\pm$0.06 & \cellcolor{2best} 0.85$\pm$0.00& \cellcolor{2best} 0.65$\pm$0.02 & \cellcolor{2best} 0.61$\pm$0.03 & \cellcolor{2best} 0.84$\pm$0.03 &  \cellcolor{2best} 0.93$\pm$0.01 & \cellcolor{2best} 0.49$\pm$0.07 &  \cellcolor{2best}0.68$\pm$0.03 & 0.50$\pm$0.04 & \cellcolor{2best} 0.68$\pm$0.14 & \cellcolor{2best} 0.65$\pm$0.03 & \cellcolor{2best} 0.53$\pm$0.03 &  \cellcolor{2best}  0.82$\pm$0.09  &   \cellcolor{2best}  0.88$\pm$0.01 & \cellcolor{2best}  0.94$\pm$0.00& \cellcolor{2best} 0.54$\pm$0.05
\\
\bottomrule
\end{tabular}
}
\label{tab:ood_robust}
\end{table*}

\subsection{Experiments}

\subsubsection{Comparison to other LL--PDL methods} To evaluate the performance of the LUR and RLUR approaches, we compare them to several LL--PDL methods. As general baselines, we consider a non-probabilistic reference, using the softmax distribution from a fine-tuned regular DNN, and a Deep Ensemble (DE) \cite{lakshminarayanan2017simple} consisting of five models. The LL--PDL methods included in our comparison are: Sub-Ensembles (SE) \cite{lee2015m,valdenegro2023sub}, Last Layer Hamiltonian Monte Carlo (LL--HMC) \cite{neal2011mcmc, hoffman2014no,vellenga2025llhmc}, Deep Deterministic Uncertainty (DDU) \cite{mukhoti2023deep}, Bayes-by-Backprop Last Layer (BBB--LL) \cite{blundell2015weight}, Packed-Ensembles Last Layer (PE--LL) \cite{laurent2023packed}, Variational Bayes Last Layer (VBLL) \cite{harrison2024variational}, and Repulsive Last Layer Ensembles (RLLE) \cite{d2021repulsive, steger2024function}. DDU uses the same (non-spectrally normalized) latent representations as the other PDL methods. 

\subsubsection{Uncertainty-based out-of-distribution detection}
The driving datasets do not, and never will, cover all possible driving maneuver types, scenarios, regions, and behaviors. Additionally, it is not realistic to evaluate across the included datasets due to differences in video quality and data collection setups. Therefore, we assess the ability of the LUR and LL--PDL methods to produce higher uncertainty estimates for OOD instances. To create realistic semantic OOD instances, we train the models without the most prevalent driving maneuver, or without the least common maneuver for each dataset\footnote{Due to the limited size of the ROAD dataset after removal of the most common maneuver,, we only consider the OOD scenario with removing the least occurring maneuvers.}. The held-out maneuvers from the training and test datasets are used as OOD instances. We compare the estimated uncertainty for both included and excluded maneuvers to assess whether the PDL methods can distinguish and identify \textit{``unknown"} maneuvers not encountered during training.


\begin{figure*}[t]
\centering
\begin{tikzpicture}
\begin{groupplot}[
    group style={
        group size=4 by 2,        
        horizontal sep=0.8cm,      
        vertical sep=0.8cm,          
        x descriptions at=edge bottom, 
        y descriptions at=edge left    
    },
    grid=major,
    width=5cm, height=2.8cm,
    ybar, 
    ymin=0, ymax=100,
    xtick=data,
    symbolic x coords={SE / LUR, BBB, PE},
    enlarge x limits=0.3,
    tick label style={font=\tiny},
    label style={font=\tiny},
        xlabel style={yshift=0.1cm},  
    y tick label style={
        /pgf/number format/fixed,
        /pgf/number format/precision=1,
        /pgf/number format/fixed zerofill=true, 
    },
    tick label style={font=\fontsize{4}{4}\selectfont},
     ylabel style={yshift=-0.10cm},
]

\nextgroupplot[
    title={\tiny{\textbf{AIDE} \citep{yang2023aide}}},
    ylabel={\tiny{F1 (\%)}},
          title style={yshift=-0.2cm},
    ymin=0, ymax=100,
              y tick label style={
        /pgf/number format/fixed,
        /pgf/number format/precision=0,
        /pgf/number format/fixed zerofill=true, 
    },
           xlabel=\empty,]

\addplot+[
    error bars/.cd,
    y dir=both,
    y explicit,
] coordinates {
    (SE / LUR,82.6)  +- (0.21,0.21)
    (BBB,83.13) +- (0.26,0.26)
    (PE,83.37) +- (0.02,0.02)
};
\addplot+[
    error bars/.cd,
    y dir=both,
    y explicit,
] coordinates {
    (SE / LUR,83.24) +- (0.18,0.18)
    (BBB,82.82) +- (0.39,0.39)
    (PE,83.35) +- (0.23,0.23)
};

\nextgroupplot[
    title={\tiny{\textbf{B4C} \citep{jain2015car}}},
     title style={yshift=-0.2cm},
    xlabel=\empty,
    ymin=0, ymax=100,
]
\addplot+[
    error bars/.cd,
    y dir=both,
    y explicit,
] coordinates {
    (SE / LUR,94.38)  +- (0.04,0.04)
    (BBB,94.28) +- (0.87,0.87)
    (PE,95.02) +- (0.02,0.02)
};
\addplot+[
    error bars/.cd,
    y dir=both,
    y explicit,
] coordinates {
    (SE / LUR,95.16) +- (0.29,0.29)
    (BBB,94.62) +- (0.29,0.29)
    (PE,95.01) +- (0.01,0.01)
};

\nextgroupplot[
    title={\tiny{\textbf{ROAD} \citep{singh2022road}}},
     ylabel=\empty,
     ymin=0, ymax=100,
          title style={yshift=-0.2cm},
           xlabel=\empty,]
\addplot+[
    error bars/.cd,
    y dir=both,
    y explicit,
] coordinates {
    (SE / LUR,45.63)  +- (2.45,2.45)
    (BBB,56.14) +- (4.37,4.37)
    (PE,59.28) +- (1.38,1.38)
};
\addplot+[
    error bars/.cd,
    y dir=both,
    y explicit,
] coordinates {
    (SE / LUR,55.82) +- (2.01,2.01)
    (BBB,41.47) +- (2.83,2.83)
    (PE,57.84) +- (1.51,1.51)
};

\nextgroupplot[
    title={\tiny{\textbf{NuScenes} \citep{caesar2020nuscenes}}},
    ylabel=\empty,
    ymin=0, ymax=100,
          title style={yshift=-0.2cm},
           xlabel=\empty,]
\addplot+[
    error bars/.cd,
    y dir=both,
    y explicit,
] coordinates {
    (SE / LUR,45.42)  +- (0.26,0.26)
    (BBB,44.70) +- (1.41,1.41)
    (PE,45.74) +- (0.1,0.1)
};
\addplot+[
    error bars/.cd,
    y dir=both,
    y explicit,
] coordinates {
    (SE / LUR,45.4) +- (0.07,0.07)
    (BBB,44.00) +- (0.73,0.73)
    (PE,45.68) +- (0.25,0.25)
};


\nextgroupplot[
    title={\empty},
    ylabel={PR-AUC},
          title style={yshift=-0.2cm},
           xlabel=\empty,
            ymin=0, ymax=1.02,]
\addplot+[
    error bars/.cd,
    y dir=both,
    y explicit,
] coordinates {
    (SE / LUR,0.21) +- (0.01,0.01)
    (BBB,0.72) +- (0.09128,0.09128)
    (PE,0.5237) +- (0.001,0.001)
};
\addplot+[
    error bars/.cd,
    y dir=both,
    y explicit,
] coordinates {
    (SE / LUR,0.97) +- (0.01,0.01)
    (BBB,0.45) +- (0.01,0.01)
    (PE,1.0) +- (0.00,0.00)
};
\nextgroupplot[
    title={\empty},
    ylabel={\empty},
          title style={yshift=-0.2cm},
           xlabel=\empty,
            ymin=0, ymax=1,]
\addplot+[
    error bars/.cd,
    y dir=both,
    y explicit,
] coordinates {
    (SE / LUR,0.54) +- (0.01,0.01)
    (BBB,0.742) +- (0.09128,0.09128)
    (PE,0.5237) +- (0.001,0.001)
};
\addplot+[
    error bars/.cd,
    y dir=both,
    y explicit,
] coordinates {
    (SE / LUR,0.93) +- (0.01,0.01)
    (BBB,0.89) +- (0.01,0.01)
    (PE,0.95) +- (0.03,0.03)
};

\nextgroupplot[
    title={\empty},
    ylabel={\empty},
          title style={yshift=-0.2cm},
           xlabel=\empty,
            ymin=0, ymax=1,]
\addplot+[
    error bars/.cd,
    y dir=both,
    y explicit,
] coordinates {
    (SE / LUR,0.33) +- (0.02,0.02)
    (BBB,0.49) +- (0.13,0.13)
    (PE,0.52) +- (0.01,0.01)
};
\addplot+[
    error bars/.cd,
    y dir=both,
    y explicit,
] coordinates {
    (SE / LUR,1.0) +- (0.00,0.00)
    (BBB,1.0) +- (0.00,0.00)
    (PE,1.0) +- (0.00,0.00)
};

\nextgroupplot[
    title={\empty},
    ylabel={\empty},
          title style={yshift=-0.2cm},
           xlabel=\empty,
            ymin=0, ymax=1,]
\addplot+[
    error bars/.cd,
    y dir=both,
    y explicit,
] coordinates {
    (SE / LUR,0.26) +- (0.01,0.01)
    (BBB,0.53) +- (0.03,0.03)
    (PE,0.52) +- (0.00,0.00)
};
\addplot+[
    error bars/.cd,
    y dir=both,
    y explicit,
] coordinates {
    (SE / LUR,0.93) +- (0.01,0.01)
    (BBB,0.98) +- (0.01,0.01)
    (PE,0.95) +- (0.02,0.02)
};
\end{groupplot}

\end{tikzpicture}
\begin{tikzpicture}
\hspace{0.5cm}
\begin{axis}[
    hide axis,
    xmin=0, xmax=1, ymin=0, ymax=1,
    legend style={at={(0.5,0.5)}, anchor=center, legend columns=3},
    legend entries={\tiny Last Layer, \tiny Latent Uncertainty Representation}
]
\addlegendimage{ybar, ybar legend, draw=none, fill=blue!60}
\addlegendimage{ybar, ybar legend, draw=none, fill=red!60}
\end{axis}
\end{tikzpicture}
    \caption{Comparison between last layer and LUR approaches in terms of in-distribution F1-score (top row) and OOD min PR-AUC performance (bottom row) across different transformation layer types. Whiskers indicate two standard errors of the mean computed over five random seeds.}
    \label{fig:translayer_type}
\end{figure*}
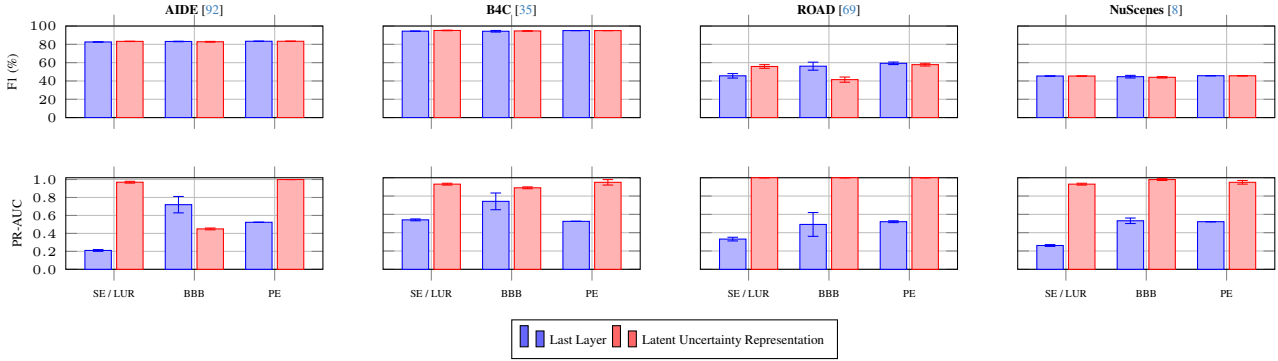

\begin{figure*}[t]
 \centering
\begin{tikzpicture}
   
\begin{groupplot}[
    group style={
        group size=4 by 4,
        horizontal sep=0.8cm,
        vertical sep=0.5cm,
    },
    width=5.00cm,
    height=2.8cm,
    grid=major,
    xlabel={Number of transformation layers},
    xtick={5,10,15,20,25,30,35,40,45,50},
    tick label style={font=\tiny},
    label style={font=\tiny},
    y tick label style={
        /pgf/number format/fixed,
        /pgf/number format/precision=1,
        /pgf/number format/fixed zerofill=true, 
    },
    tick label style={font=\fontsize{4}{4}\selectfont},
    ylabel style={yshift=-0.10cm},  
    xlabel style={yshift=0.15cm},  
]

\nextgroupplot[
    title={\tiny{\textbf{AIDE} \citep{yang2023aide}}},
    ylabel={\tiny{F1 (\%)}},
    title style={yshift=-0.2cm},
    y tick label style={
        /pgf/number format/fixed,
        /pgf/number format/precision=0,
        /pgf/number format/fixed zerofill=true, 
    },
     xlabel=\empty,
    ymin=70,
    ymax=90 
]

\addplot[mark=square*, thin, line width=0.25pt, mark options={fill=pink},  mark size=1.5pt] coordinates {
(5,82.047)
(10,82.361)
(15,83.27)
(20,82.586)
(25,81.489)
(30,82.545)
(35,81.9)
(40,81.963)
(45,82.846)
(50,81.854)
};

\addplot[mark=diamond*, thin, line width=0.25pt, mark options={fill=green}, mark size=1.5pt] coordinates {
(5, 81.271)
(10,82.834)
(15,82.90)
(20,81.406)
(25,83.114)
(30,81.92)
(35,82.374)
(40,82.061)
(45,82.539)
(50,82.459)
            };

  \addplot[mark=pentagon*, thin, line width=0.25pt, mark options={fill=red}, mark size=1.5pt] coordinates {
(5,81.213)
(10,81.872)
(15,76.251)
(20,79.691)
(25,80.413)
(30,79.935)
(35,77.753)
(40,81.804)
(45,78.587)
(50,79.117)
  };

\nextgroupplot[
    title={\tiny{\textbf{B4C} \citep{jain2015car}}},
          title style={yshift=-0.2cm},
           xlabel=\empty,
       y tick label style={
        /pgf/number format/fixed,
        /pgf/number format/precision=0,
        /pgf/number format/fixed zerofill=true, 
    },
    ymin=50,
    ymax=100 
]
\addplot[mark=square*, thin, line width=0.25pt, mark options={fill=pink},  mark size=1.5pt] coordinates {
(5,90.025)
(10,95.003)
(15,92.09)
(20,92.338)
(25,94.167)
(30,92.834)
(35,93.733)
(40,93.547)
(45,94.382)
(50,94.382)
};
    \addplot[mark=diamond*, thin, line width=0.25pt, mark options={fill=green}, mark size=1.5pt] coordinates {
(5,91.063)
(10,94.111)
(15,91.763)
(20,90.237)
(25,95.003)
(30,92.217)
(35,94.382)
(40,92.658)
(45,93.715)
(50,92.595)
            };
  \addplot[mark=pentagon*, thin, line width=0.25pt, mark options={fill=red}, mark size=1.5pt] coordinates {
(5,81.962)
(10,79.969)
(15,76.898)
(20,57.25)
(25,71.508)
(30,92.864)
(35,90.317)
(40,87.373)
(45,90.465)
(50,91.564) 
  };

            \nextgroupplot[
    title={\tiny{\textbf{ROAD} \citep{singh2022road}}},
          title style={yshift=-0.2cm},
    y tick label style={
        /pgf/number format/fixed,
        /pgf/number format/precision=0,
        /pgf/number format/fixed zerofill=true, 
    },
     xlabel=\empty,
    ymin=10,
    ymax=65 
]
  \addplot[mark=square*, thin, line width=0.25pt, mark options={fill=pink},  mark size=1.5pt] coordinates {
  (5,41.344)
(10,42.443)
(15,46.153)
(20,41.614)
(25,31.235)
(30,38.99)
(35,35.382)
(40,38.006)
(45,39.851)
(50,37.802)
  };

    \addplot[mark=diamond*, thin, line width=0.25pt, mark options={fill=green}, mark size=1.5pt] coordinates {
(5,33.685)
(10,36.516)
(15,37.575)
(20,42.314)
(25,39.11)
(30,44.683)
(35,38.749)
(40,38.165)
(45,34.725)
(50,38.165)
            };

\addplot[mark=pentagon*, thin, line width=0.25pt, mark options={fill=red}, mark size=1.5pt] coordinates {
(5,18.862)
(10,33.256)
(15,42.388)
(20,29.936)
(25,30.277)
(30,30.678)
(35,26.326)
(40,32.219)
(45,31.28)
(50,27.35) 
  };
            \nextgroupplot[
    title={\tiny{\textbf{NuScenes} \citep{caesar2020nuscenes}}},
          title style={yshift=-0.2cm},
           xlabel=\empty,
    y tick label style={
        /pgf/number format/fixed,
        /pgf/number format/precision=0,
        /pgf/number format/fixed zerofill=true, 
    },
    ymin=20,
    ymax=50 
]

\addplot[mark=square*, thin, line width=0.25pt, mark options={fill=pink},  mark size=1.5pt] coordinates {
(5,44.493)
(10,43.324)
(15,43.547)
(20,41.548)
(25,43.05)
(30,43.487)
(35,44.109)
(40,43.313)
(45,44.422)
(50,46.572)
};

    \addplot[mark=diamond*, thin, line width=0.25pt, mark options={fill=green}, mark size=1.5pt] coordinates {
(5,33.685)
(10,36.516)
(15,37.575)
(20,42.314)
(25,39.11)
(30,44.683)
(35,38.749)
(40,38.165)
(45,34.725)
(50,38.165)
            };
  \addplot[mark=pentagon*, thin, line width=0.25pt, mark options={fill=red}, mark size=1.5pt] coordinates {
  (5,23.802)
(10,28.497)
(15,31.282)
(20,33.121)
(25,37.86)
(30,45.783)
(35,33.817)
(40,32.678)
(45,33.675)
(50,33.968)
  };

\nextgroupplot[
    ylabel={\tiny{PR-AUC}},
          title style={yshift=-0.15cm},
    ymin=0.00,
    ymax=1.00 
]

\addplot[mark=square*, thin, line width=0.25pt, mark options={fill=pink},  mark size=1.5pt] coordinates {
(5,0.331)
(10,0.432)
(15,0.551)
(20,0.582)
(25,0.568)
(30,0.5)
(35,0.592)
(40,0.592)
(45,0.623)
(50,0.536)
};

\addplot[mark=diamond*, thin, line width=0.25pt, mark options={fill=green}, mark size=1.5pt] coordinates {
(5,0.444)
(10,0.652)
(15,0.535)
(20,0.606)
(25,0.661)
(30,0.635)
(35,0.664)
(40,0.739)
(45,0.713)
(50,0.712)
            };

  \addplot[mark=pentagon*, thin, line width=0.25pt, mark options={fill=red}, mark size=1.5pt] coordinates {
(5,0.38904)
(10,0.40507)
(15,0.49263)
(20,0.51217)
(25,0.49417)
(30,0.53117)
(35,0.54257)
(40,0.52824)
(45,0.4995)
(50,0.50123)
  };

\nextgroupplot[
          title style={yshift=-0.2cm},
    ymin=0.00,
    ymax=1.00 
]
\addplot[mark=square*, thin, line width=0.25pt, mark options={fill=pink},  mark size=1.5pt] coordinates {
(5,0.637)
(10,0.684)
(15,0.683)
(20,0.71)
(25,0.696)
(30,0.692)
(35,0.727)
(40,0.709)
(45,0.709)
(50,0.757)
};
    \addplot[mark=diamond*, thin, line width=0.25pt, mark options={fill=green}, mark size=1.5pt] coordinates {
(5,0.592)
(10,0.676)
(15,0.678)
(20,0.749)
(25,0.744)
(30,0.709)
(35,0.713)
(40,0.703)
(45,0.728)
(50,0.747)
            };
  \addplot[mark=pentagon*, thin, line width=0.25pt, mark options={fill=red}, mark size=1.5pt] coordinates {
(5,0.65345)
(10,0.71843)
(15,0.67539)
(20,0.69638)
(25,0.69311)
(30,0.74452)
(35,0.69929)
(40,0.75422)
(45,0.74898)
(50,0.74317)
  };

            \nextgroupplot[
          title style={yshift=-0.2cm},
    ymin=0.00,
    ymax=1.00 
]
  \addplot[mark=square*, thin, line width=0.25pt, mark options={fill=pink},  mark size=1.5pt] coordinates {
 (5,0.44574)
(10,0.62914)
(15,0.69561)
(20,0.84332)
(25,0.74015)
(30,0.82843)
(35,0.71979)
(40,0.75771)
(45,0.75147)
(50,0.73715)
  };

    \addplot[mark=diamond*, thin, line width=0.25pt, mark options={fill=green}, mark size=1.5pt] coordinates {
(5,0.508)
(10,0.595)
(15,0.56)
(20,0.679)
(25,0.684)
(30,0.724)
(35,0.632)
(40,0.836)
(45,0.738)
(50,0.684)
            };

\addplot[mark=pentagon*, thin, line width=0.25pt, mark options={fill=red}, mark size=1.5pt] coordinates {
(5,0.51389)
(10,0.62341)
(15,0.61384)
(20,0.7827)
(25,0.7356)
(30,0.77442)
(35,0.70624)
(40,0.80308)
(45,0.72982)
(50,0.77043)
  };
            \nextgroupplot[
          title style={yshift=-0.2cm},
    ymin=0.00,
    ymax=1.00
]

\addplot[mark=square*, thin, line width=0.25pt, mark options={fill=pink},  mark size=1.5pt] coordinates {
(5,0.57814)
(10,0.60696)
(15,0.66986)
(20,0.6953)
(25,0.70834)
(30,0.69404)
(35,0.69057)
(40,0.73688)
(45,0.70096)
(50,0.73226)
};

    \addplot[mark=diamond*, thin, line width=0.25pt, mark options={fill=green}, mark size=1.5pt] coordinates {
(5,0.58119)
(10,0.67323)
(15,0.6133)
(20,0.72949)
(25,0.68316)
(30,0.69285)
(35,0.68475)
(40,0.70027)
(45,0.71699)
(50,0.70653)
            };
  \addplot[mark=pentagon*, thin, line width=0.25pt, mark options={fill=red}, mark size=1.5pt] coordinates {
(5,0.64215)
(10,0.67394)
(15,0.66725)
(20,0.70697)
(25,0.71688)
(30,0.7429)
(35,0.72928)
(40,0.68087)
(45,0.72643)
(50,0.70336)
  };
        \end{groupplot}
          
    \end{tikzpicture}
    
\begin{tikzpicture}
\hspace{0.75cm}
\begin{axis}[
    hide axis,
    xmin=0, xmax=1, ymin=0, ymax=1,
    legend style={at={(0.5,0.5)}, anchor=center, legend columns=4},
    legend entries={\tiny{KDE},\tiny{SGE}, \tiny{SSGE}}
]

\addlegendimage{mark=pentagon*, thin, line width=0.25pt, mark options={fill=red}, mark size=2.0pt}
\addlegendimage{mark=square*, thin, line width=0.25pt, mark options={fill=pink}, mark size=2pt}
\addlegendimage{mark=diamond*, thin, line width=0.25pt, mark options={fill=green}, mark size=2.0pt}

\end{axis}
\end{tikzpicture}
    \caption{In-distribution classification performance (top row) and OOD min detection performance (bottom row) for different kernel types and numbers of transformation layers in the RLUR approach for a single random seed.}
    \label{fig:kernel_overview_per}
\end{figure*}
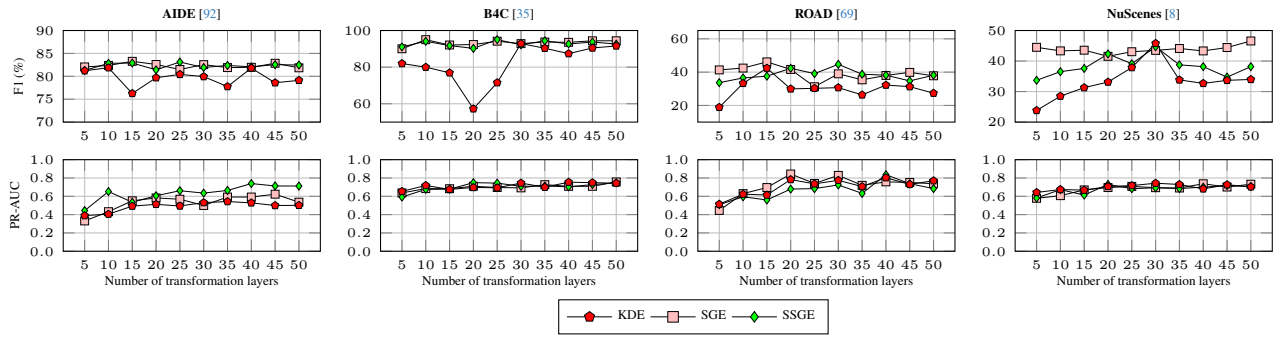

\section{Results}
To ensure the robustness of our findings in both the in-distribution and OOD scenarios, we conducted grid searches using five random seeds. This approach allows us to assess variability across repeated trials and more accurately reflect the potential outcomes of re-running the experiments under different initialization conditions \cite{bouthillier2019unreproducible,vellenga2025llhmc}.

 \subsection{Action and intention recognition performance}
Table \ref{tab:id_perf} presents the average performance of the top-performing hyperparameters for each random seed. Compared to the other LL--PDL methods, we observe that the LUR and RLUR approaches occasionally match the top-performing PDL methods, but do not consistently outperform them. For example, LUR yields the best F1-score for the B4C dataset and the third highest for the other datasets. RLUR only performs worse on the smaller ROAD dataset (similar to RLLE).  

\subsection{Out-of-distribution detection}
Table \ref{tab:ood_perf} reports the average OOD detection results for the top-performing hyperparameter configurations across five random seeds. In contrast to the in-distribution results, LUR consistently matches or outperforms the other included PDL methods. Alongside LUR, RLLE also demonstrates top performance across all scenarios. The top-performing hyperparameter configurations of the RLUR approach match the OOD detection performance of LL--HMC, but are slightly worse compared to LUR and RLLE. 


\paragraph{\textbf{Robustness across random seeds.}} Table \ref{tab:ood_robust} presents the best average uncertainty-based OOD detection performance per hyperparameter configuration across the five random seeds. For the LUR approach, the OOD detection performance (especially for the OOD min scenarios) worsens compared to the average top performing hyperparameter configuration across random seeds. For the repulsive approaches, the RLLE performance achieves OOD detection results that are closest to the average top-performing hyperparameter configurations in Table \ref{tab:ood_perf}. For the RLUR approach, the performance is lower across the OOD scenarios, but remains considerably better than the other included PDL methods. This suggests that the best-performing hyperparameter configurations of the repulsive approaches are more robust across random seeds.

\paragraph{\textbf{Different types of transformation layers.}} The type of layer responsible for producing the additional latent representations is not fixed. Therefore, it is possible to substitute the linear transformation layers with other PDL methods (\eg, Bayes-by-Backprop \cite{blundell2015weight} or a packed-ensemble \cite{laurent2023packed}). Figure \ref{fig:translayer_type} illustrates the comparison between the last layer and LUR-based approaches for both in-distribution classification and OOD min performance. Across the datasets and five random seeds, we observe little impact on the in-distribution DAR and DIR performance. However, from an OOD detection perspective, we observe that the LUR approaches significantly outperform the equivalent last layer approaches, except for BBB--LL on the AIDE dataset. This suggests that adding transformation layers, regardless of the underlying layer type, can be effective for detecting OOD instances.


\paragraph{\textbf{Does the kernel type matter?}} As alternatives to the SSGE kernel, D'Angelo and Fortuin \cite{d2021repulsive} describe two other kernel types: Kernel Density Estimation (KDE) with a radial basis function, and the Stein Gradient Estimator (SGE) \cite{li2018gradient}. Intuitively, KDE considers only how each individual particle (transformation layer) interacts with the others. SGE improves upon this by also accounting for interactions among all particles, providing a more balanced view. 

To understand the effect of kernel type on RLUR, we evaluated the influence of different numbers of transformation layers on in-distribution and OOD min performance. The first rows of \Cref{fig:kernel_overview_per} present the F1-score, and the second rows present the PR-AUC for the top-performing hyperparameter configurations for a single random seed. The SGE and SSGE kernels exhibit stable performance for both in-distribution and OOD detection across datasets and varying numbers of transformation layers (particles). However, the results also suggest that for each dataset, there is a KDE configuration that matches the performance of the other kernels.


\begin{table}[t]
\centering
\caption{Variance-based latent representation OOD detection.}
\resizebox{0.48\textwidth}{!}{
\begin{tabular}{lc|c|c|c||c|c|c}
\toprule
\multicolumn{2}{c|}{} & \multicolumn{3}{c||}{\textbf{LUR}} & \multicolumn{3}{c}{\textbf{RLUR}} \\
\cmidrule{3-8}


& & \begin{tabular}{@{}c@{}}\textbf{ROC-}\\ \textbf{AUC} ($\uparrow$)\end{tabular} & \begin{tabular}{@{}c@{}}\textbf{PR-}\\\textbf{AUC} ($\uparrow$)\end{tabular} & \begin{tabular}{@{}c@{}}\textbf{FPR95}\\($\downarrow$)\end{tabular} & \begin{tabular}{@{}c@{}}\textbf{ROC-}\\ \textbf{AUC} ($\uparrow$)\end{tabular} & \begin{tabular}{@{}c@{}}\textbf{PR-}\\\textbf{AUC} ($\uparrow$)\end{tabular} & \begin{tabular}{@{}c@{}}\textbf{FPR95}\\($\downarrow$)\end{tabular} \\ \midrule

\multirow{2}{*}{\textbf{AIDE}} 
& \small{OOD min} & 0.69 & 0.23 & 0.69 & 0.55 & 0.19 & 0.92 \\
& \small{OOD max} & 0.66 & 0.12 & 0.99 & 0.67 & 0.15 & 0.99 \\ 
\midrule

\multirow{2}{*}{\textbf{B4C}} 
& \small{OOD min} & 0.36 & 0.26 & 0.98 & 0.48 & 0.31 & 0.98 \\
& \small{OOD max} & 0.32 & 0.65 & 0.94 & 0.87 & 0.67 & 0.91 \\ 
\midrule

\multirow{1}{*}{\textbf{ROAD}} 
& \small{OOD min} & 0.85 & 0.84 & 0.67 & 0.40 & 0.31 & 1.00 \\  \midrule

\multirow{2}{*}{\textbf{NuScenes}} 
& \small{OOD min} & 0.46 & 0.17 & 0.94 & 0.51 & 0.19 & 0.82 \\
& \small{OOD max} & 0.60 & 0.35 & 0.98 & 0.67 & 0.47 & 0.92 \\ 

\bottomrule
\end{tabular}
}
\label{tab:variance}
\end{table}

\paragraph{\textbf{Variance-based OOD detection results.}} Instead of producing additional predictions, the LUR and RLUR approaches produce additional latent representations. Therefore, it is also possible to quantify the variance across these latent representations instead of using predictive entropy to measure the uncertainty. Table \ref{tab:variance} presents the variance-based OOD detection results for the top-performing LUR and RLUR hyperparameter configurations for a single random seed. Compared to predictive entropy-based results in Table \ref{tab:ood_perf}, the variance-based latent representation uncertainty estimates for both LUR and RLUR models are significantly less effective at detecting OOD instances.


\paragraph{\textbf{Computational impact.}} From a training perspective, LUR only requires additional transformation layers and a modified loss function. The repulsive approaches are more computationally demanding due to the comparisons between the particles (transformation layers). The computational cost to produce additional predictions is higher for (R)LUR compared to a last layer approach. For example, the ViT-Base architecture \cite{tong2022videomae} requires approximately $361 \times 10^9$ floating point operations (FLOPs) to process a 16-frame video with a resolution of 224 by 224. Each additional last layer with, for example, five classes, requires $7.68 \times 10^3$ FLOPs, whereas for LUR, an additional $1.18 \times 10^6$ FLOPs are required per additional latent representation next to the additional last layer computations. On a relative scale, the added FLOPs can be considered negligible, but LUR does require more computation compared to a last layer approach to produce uncertainty estimations. 

\section{Discussion \& Conclusion}
Deploying artificial intelligence (AI) in real-world safety-critical applications requires a cautious approach. This is especially important in environments with limited computational resources, shared workloads across tasks, and reliance on pre-trained or shared latent representations for downstream tasks. To support efficient uncertainty estimation in these contexts, we evaluated two approaches based on additional latent representations. We evaluated the latent uncertainty representation (LUR) and its repulsive variant (RLUR) on two open-source video driver action recognition (DAR) datasets and two driver intention recognition (DIR) datasets. We also release over 28,000 frame-level DAR labels and 1,194 video-level DIR labels for the NuScenes dataset \cite{caesar2020nuscenes}.

In terms of in-distribution classification performance, the last layer and LUR-based approaches achieve comparable performance. From an uncertainty-based out-of-distribution (OOD) detection perspective, LUR matches the top-performing last layer approaches, despite its simplicity and lack of explicit diversity. Future work could investigate the effects of deeper or alternative transformation layers to produce additional latent representations, or extend the loss function with a reconstruction term, as in \cite{upadhyay2023probvlm}. Additionally, the influence of alternative uncertainty estimation measures or approaches (\eg, absolute distance-based \cite{durasov2024zigzag}, order-consistent split \cite{haas2025aleatoric}) could be considered.

For driving automation applications, it is common for weather and lighting conditions to change, for sensor observations to occasionally be blurry, or for other forms of noise to be introduced during data processing. Future research could evaluate the robustness of LUR or last layer PDL approaches under such corruptions, for instance by applying the benchmarks proposed in \citet{michaelis2019benchmarking}. However, not all synthetic corruptions accurately reflect real-world behavior. For instance, a change in weather may influence driver caution in ways that artificial perturbations cannot capture. Alternatively, simulated data could be used to introduce variation in the poses of nearby road users, as long as these changes do not alter driver actions or intentions \cite{tonderski2024neurad}.


{
    \small
    \bibliographystyle{ieeenat_fullname}
    \bibliography{main}
}

\clearpage

\appendices
\section*{APPENDICES}

\section{NuScenes -- Annotation Criterion} \label{app:annotation}
Given the lack of ego-vehicle action and intention labels in open-source naturalistic driving datasets \cite{vellenga2022driver}, we annotated NuScenes \cite{caesar2020nuscenes}, one of the most popular open-source traffic scene datasets. This dataset was collected in Boston and Singapore, and consists of driving scenarios recorded at different times of day and under varying weather conditions. Compared to other datasets with driving action or intention labels (\eg, \cite{jain2015car,ramanishka2018toward,singh2022road}), NuScenes provides high-resolution exterior camera footage, is easy to access, and enables future multi-modal experiments.

\begin{figure}[t]
    \centering
    \begin{subfigure}[b]{0.6\columnwidth}
        \centering
         \includegraphics[trim={0cm 9.2cm 22cm 0},clip,  width=\columnwidth, keepaspectratio]{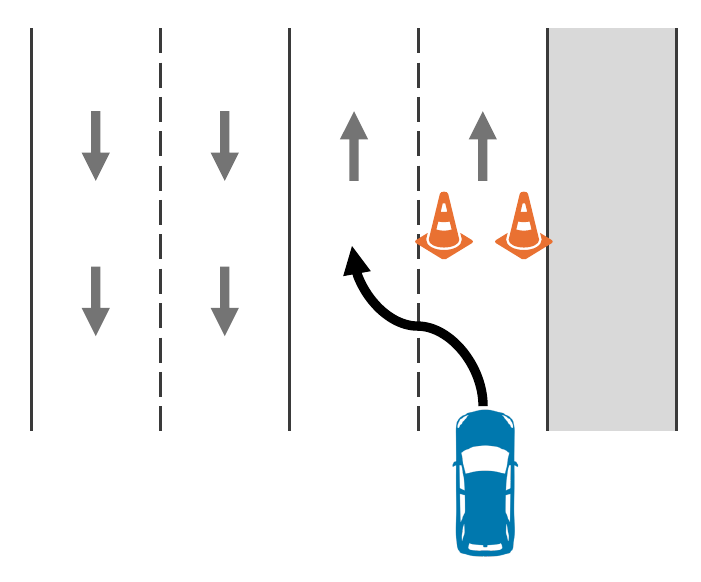}
        \caption{Forced lane change due to road construction.}
        \label{fig:roadwork}
    \end{subfigure}
    \vspace{0.25cm}
    
     \begin{subfigure}[b]{0.6\columnwidth}
        \centering
         \includegraphics[trim={0cm 9.2cm 22cm 0},clip,  width=\columnwidth, keepaspectratio]{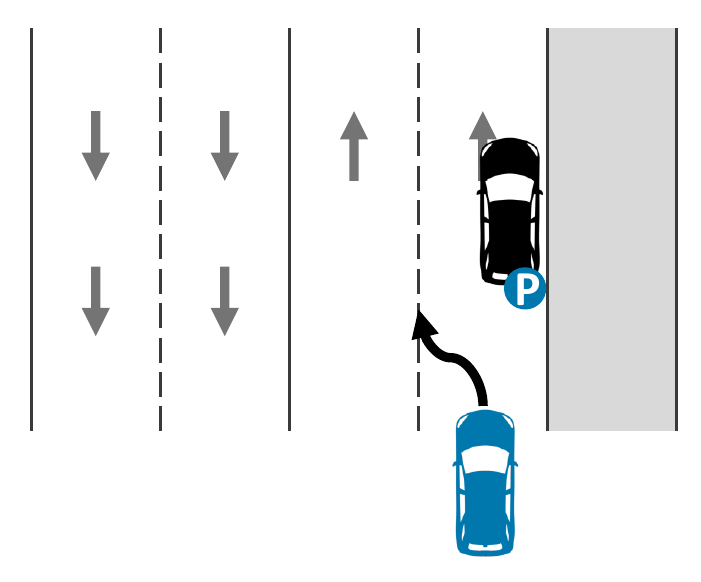}
        \caption{Parked car forcing a lane change.}
        \label{fig:parkedcar}
    \end{subfigure}
    \vspace{0.25cm}
    
     \begin{subfigure}[b]{0.6\columnwidth}
        \centering
        \centering
         \includegraphics[trim={0cm 9cm 25cm 0},clip,  height=5cm, keepaspectratio]{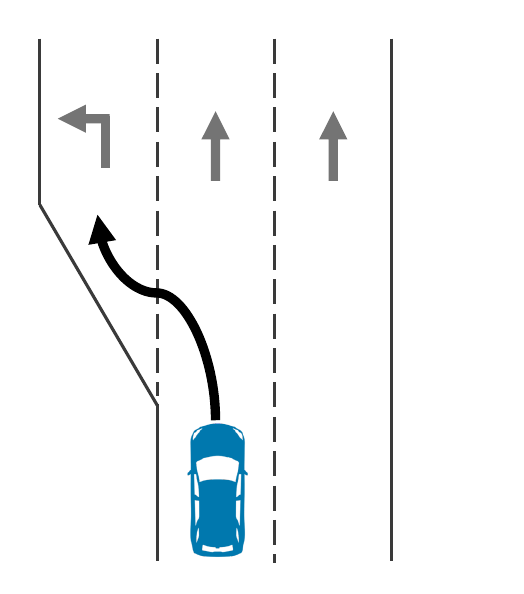}
        \caption{Pre-turn maneuver lane change.}
        \label{fig:preturn}
    \end{subfigure}
  
     \caption{Urban lane change examples.}
    \vspace{-0.5cm} 
    
    \end{figure}

\begin{figure}[t]
  \centering
  \includegraphics[trim={0cm 9.5cm 25cm 0},clip,  height=6cm, keepaspectratio]{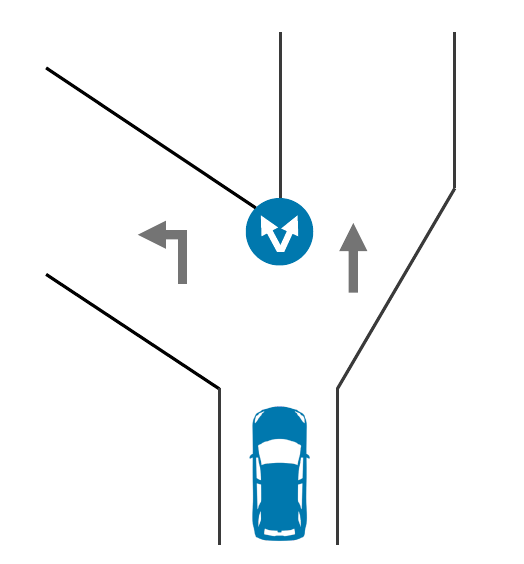}
  \caption{Schematic overview of a choice driving scenario.}
  \label{fig:choice}
\end{figure}

Driving in an urban environment is often interrupted by waiting for traffic lights or other road users before completing maneuvers.Similar to Singh \etal \cite{singh2022road}, we include \textit{waiting} as one of the actions. To avoid ambiguous or extremely rare driving maneuvers, we exclude \textit{u-turns}, \textit{move-left}, and \textit{move-right} \cite{ramanishka2018toward,singh2022road}. We treat the scenarios shown in Figures \ref{fig:roadwork}, \ref{fig:parkedcar}, and \ref{fig:preturn} as lane change maneuvers, because they require the driver to perform similar procedures (checking side mirrors and blind spots) before changing lanes. When the driver must choose between proceeding straight or turning (see Figure \ref{fig:choice}), we consider the turn maneuver to begin at the moment a change in direction is observed in the video frames. Consistent with previous ego-vehicle driving maneuver benchmarks, we include the following maneuvers: \textit{moving forward}, \textit{turn left}, \textit{turn right}, \textit{left lane change}, \textit{right lane change}, and \textit{waiting}.

\begin{figure*}[t]
  \centering
  \hspace*{-0.0cm}
  \includegraphics[trim={0cm 13cm 0cm 0},clip,  width=0.95\textwidth, keepaspectratio]{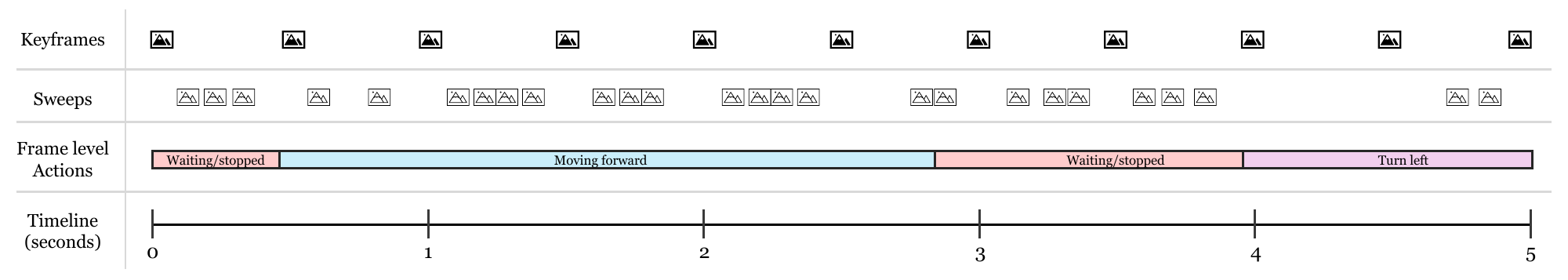}
  \vspace{-0.3cm}
  \caption{Schematic overview of a driving scene denoting the inconsistency in available frames, and the annotations.}
    \vspace{-0.0cm}
  \label{fig:annotation}
\end{figure*}

Frames in the NuScenes \cite{caesar2020nuscenes} dataset are collected continuously from six cameras operating at 12 Hz, as well as from radar and a LiDAR system. In some cases, the gap between frames varied due to high system load, which can also occur in real-world driving scenarios. However, to avoid missing entire maneuvers (such as a turn), we only include sequences that span at least three seconds (see Figure \ref{fig:annotation} for a schematic overview). Annotating ego-vehicle driving maneuvers can be subjective. For example, it is not always clear at what point a turn or lane change maneuver is considered finished. To ensure consistency, we follow a similar approach to Kung \etal \cite{kung2024action}, with a single annotator responsible for all labels. Furthermore, we apply the following annotation criteria to ensure consistency: for each sequence, we annotate the start and end frames with a driving maneuver label. We then identify the frame where a new maneuver begins; all frames before this switch point inherit the previous driving maneuver label.

Drivers can possess multiple driving intentions, but we can only verify a driver's intention to perform a maneuver if the maneuver is actually observed. While this introduces a hindsight bias, it is necessary to balance which intentions are both relevant and feasible to recognize.



\clearpage

       \begin{algorithm}[t]
        \caption{PyTorch (R)LUR model wrapper example}
        \begin{lstlisting}[language=Python, numbers=none]
from torch import nn
from copy import deepcopy 

class LUR_wrapper(nn.Module):
    def __init__(self, 
                 model:nn.Module,
                 model_dim:int,
                 num_representations:int):
        super().__init__()
        # Split the encoder and classification layer of the original model
        self.encoder = model
        self.classifier = deepcopy(model.classifier)
        self.encoder.classifier = nn.Identity()
        self.encoder = self.encoder.eval()
        
        # Initialize the transformation layers
        self.num_representations = num_representations
        self.transformation_layers = [nn.Linear(model_dim, model_dim) for _ in range(self.num_representations)]

    def forward(self, x):
        # Extract latent feature representation from raw input data with an encoder 
        z = self.encoder(x)
        # Store the representation in a list 
        z_reps = [z]
        # Process the latent representation through the trainable transformation layers
        for i in range(self.num_representations):
            z_trans = self.transform_layers[i](z)
            # Store the projected representation 
            z_reps.append(z_trans) 
        # Predict for each representation 
        y = [self.classifier(z) for z in z_reps]
        return y
    \end{lstlisting}
        \label{alg:pseudo-code}
        \end{algorithm}

\section{Pseudo Torch code} \Cref{alg:pseudo-code} presents the PyTorch pseudo-code for the (R)LUR inference. When instantiating the model, the LUR wrapper expects the original pre-trained or fine-tuned model, the latent representation dimension, and the number of representations. From the original model, the classifier is removed to extract the latent representations. The number of representations defines how many transformation layers are initialized. To produce the set of predictions, as shown in \Cref{fig:LUR_schema}, the forward def expects the raw input data $x$. Then the data is first encoded into the latent representation $z$. Next, the original latent representation is stored in a list and then processed through the transformation layers. The original latent representation $z$ and the transformed representations $z_{trans}^{(i)}$ are passed through the same classification layer, resulting in multiple predictions $\hat{y}$.

\clearpage

\section{Best performing average hyperparameter configuration across random seeds} \label{app:robust}  
Consistent with the findings of \cite{vellenga2025llhmc}, we observed that the best average hyperparameter configuration across seeds yielded significantly worse performance. Table \ref{tab:avg_across_seeds} presents the average in-distribution performance and two standard errors of the mean for the best average hyperparameter configurations across five random seeds. Compared to Table \ref{tab:id_perf}, the performance is slightly worse for the AIDE and NuScenes datasets, but significantly worse for the B4C and ROAD datasets.

\begin{table*}[t]
\centering
\caption{Average in-distribution grid search results and two standard errors of the mean for the best average hyperparameter configuration across five random seeds. Acc=Accuracy, ACE=Adaptive Calibration Error, rAULC=relative Area under the lifted curve.}
\vspace{-0.10cm}
\resizebox{0.98\linewidth}{!}{
\begin{tabular}{l|c|c|c|c||c|l|c|c||c|c|c|c||c|c|c|c}
\toprule
 & \multicolumn{4}{c||}{\textbf{AIDE} \cite{yang2023aide}} & \multicolumn{4}{c||}{\textbf{B4C} \cite{jain2015car}} & \multicolumn{4}{c||}{\textbf{ROAD} \cite{singh2022road}}  & \multicolumn{4}{c}{\textbf{NuScenes} \cite{caesar2020nuscenes}}  \\
 & \textbf{Acc ($\uparrow$)} & \textbf{F1 ($\uparrow$)} & \textbf{ACE ($\downarrow$)} & \textbf{rAULC ($\uparrow$)} & \textbf{Acc ($\uparrow$)} &  \multicolumn{1}{|c|}{\textbf{F1 ($\uparrow$)}} & \textbf{ACE ($\downarrow$)} & \textbf{rAULC ($\uparrow$)} & \textbf{Acc ($\uparrow$)} & \textbf{F1 ($\uparrow$)} & \textbf{ACE ($\downarrow$)} & \textbf{rAULC ($\uparrow$)} & \textbf{Acc ($\uparrow$)} & \textbf{F1 ($\uparrow$)} & \textbf{ACE ($\downarrow$)} & \textbf{rAULC ($\uparrow$)} \\

\midrule

\textbf{BBB--LL} &  88.47$\pm$0.41 & 82.49$\pm$0.41 & 0.030$\pm$0.001 &0.70$\pm$0.05 & 90.09$\pm$1.16 &  91.26$\pm$1.70 & 0.028$\pm$0.005 & 0.80$\pm$0.02    & 63.58$\pm$5.05 &  34.51$\pm$1.72 & 0.077$\pm$0.007 & 0.62$\pm$0.04 & 57.88$\pm$1.09 & 42.28$\pm$2.20 & 0.12$\pm$0.00& 0.57$\pm$0.02 \\

\textbf{PE--LL} & 87.85$\pm$0.63 & 80.61$\pm$2.12 &0.034$\pm$0.004 & 0.79$\pm$0.02 & 91.97$\pm$0.42 & 91.99$\pm$0.45 &    0.021$\pm$0.002 &  0.84$\pm$0.01  &  65.37$\pm$1.12 & 29.72$\pm$1.12 & 0.092$\pm$0.002 & 0.67$\pm$0.02 &60.10$\pm$0.68 & 43.10$\pm$1.04 & 0.05$\pm$0.00& 0.63$\pm$0.01 \\

\textbf{SE} & 88.11$\pm$0.13 &  81.61$\pm$0.11 &   0.028$\pm$0.001 & 0.77$\pm$0.01 &   92.14$\pm$0.34 &92.21$\pm$0.26 & 0.022$\pm$0.002 & 0.82$\pm$0.00 & 65.97$\pm$0.60 &  29.45$\pm$0.32 & 0.103$\pm$0.002 & 0.67$\pm$0.02 & 60.57$\pm$0.84 & 43.92$\pm$1.39 & 0.05$\pm$0.00& 0.62$\pm$0.02 \\

\textbf{VBLL} & 87.95$\pm$0.17 & 78.62$\pm$3.17 & 0.034$\pm$0.003 &  0.78$\pm$0.02 & 86.32$\pm$3.01 & 84.92$\pm$3.17 & 0.081$\pm$0.007 & 0.80$\pm$0.07 & 66.27$\pm$1.52 &31.29$\pm$1.71 & 0.108$\pm$0.005 &   0.63$\pm$0.03 & 60.00$\pm$0.35 & 43.79$\pm$0.60 & 0.06$\pm$0.00& 0.61$\pm$0.02  \\

\textbf{LL--HMC} &   88.44$\pm$0.22 &  82.69$\pm$0.26 &0.028$\pm$0.000 &  0.76$\pm$0.01 & 90.26$\pm$1.39 & 90.17$\pm$2.46 &  0.027$\pm$0.004 &  0.85$\pm$0.03 & 61.79$\pm$4.07 &  32.03$\pm$4.00 &   0.083$\pm$0.010 &  0.63$\pm$0.15 & 59.48$\pm$0.76 & 43.48$\pm$1.09 & 0.06$\pm$0.00& 0.63$\pm$0.01 \\

\textbf{RLLE} & 88.24$\pm$0.25 & 80.62$\pm$0.59 & 0.210$\pm$0.001& 0.61$\pm$0.09 & 82.91$\pm$1.62 & 79.59$\pm$2.20 & 0.230$\pm$0.011 & 0.62$\pm$0.04& 60.9$\pm$4.57 & 28.14$\pm$6.33 & 0.160$\pm$0.010 & 0.48$\pm$0.23 & 60.31$\pm$0.31 & 36.42$\pm$0.92 & 0.14$\pm$0.00& 0.61$\pm$0.04  \\

\textbf{LUR} & 87.98$\pm$0.41 & 81.19$\pm$0.39 & 0.030$\pm$0.001& 0.78$\pm$0.01 & 90.43$\pm$0.64 & 91.68$\pm$0.74 & 0.040$\pm$0.010 & 0.84$\pm$0.05 & 66.27$\pm$1.52 & 32.21$\pm$1.28 & 0.070$\pm$0.001& 0.63$\pm$0.02 & 59.84$\pm$0.37 & 44.23$\pm$0.39 & 0.08$\pm$0.00& 0.59$\pm$0.01 \\

\textbf{RLUR} &87.88$\pm$0.12 & 81.61$\pm$0.29 & 0.030$\pm$0.001& 0.76$\pm$0.02 & 90.77$\pm$1.74 & 91.23$\pm$2.31 & 0.030$\pm$0.010 & 0.78$\pm$0.03 & 64.18$\pm$2.99 & 35.16$\pm$1.19 & 0.080$\pm$0.012 & 0.59$\pm$0.04 &  58.24$\pm$0.47 & 42.51$\pm$0.64 & 0.12$\pm$0.00& 0.55$\pm$0.01 \\

\bottomrule
\end{tabular}
}
\label{tab:avg_across_seeds}
\end{table*}

\begin{table*}[t]
\centering
\caption{Video-based driver action and intention recognition performance for the AIDE, B4C and ROAD datasets. Each PDL method uses the latent representations produced by the \textit{Regular} model.}
\vspace{-0.10cm}
\resizebox{0.98\linewidth}{!}{
\begin{tabular}{l|c|c|c|c||c|l|c|c||c|c|c|c||c|c|c|c}
\toprule
 & \multicolumn{4}{c||}{\textbf{AIDE} \cite{yang2023aide}} & \multicolumn{4}{c||}{\textbf{B4C} \cite{jain2015car}} & \multicolumn{4}{c||}{\textbf{ROAD} \cite{singh2022road}}  & \multicolumn{4}{c}{\textbf{NuScenes} \cite{caesar2020nuscenes}}  \\
 & \textbf{Acc ($\uparrow$)} & \textbf{F1 ($\uparrow$)} & \textbf{ACE ($\downarrow$)} & \textbf{rAULC ($\uparrow$)} & \textbf{Acc ($\uparrow$)} &  \multicolumn{1}{|c|}{\textbf{F1 ($\uparrow$)}} & \textbf{ACE ($\downarrow$)} & \textbf{rAULC ($\uparrow$)} & \textbf{Acc ($\uparrow$)} & \textbf{F1 ($\uparrow$)} & \textbf{ACE ($\downarrow$)} & \textbf{rAULC ($\uparrow$)} & \textbf{Acc ($\uparrow$)} & \textbf{F1 ($\uparrow$)} & \textbf{ACE ($\downarrow$)} & \textbf{rAULC ($\uparrow$)} \\
\midrule

\textbf{SE} & 88.31$\pm$0.19 &  82.69$\pm$0.21 &  0.029$\pm$0.000 & 0.75$\pm$0.01 & 93.16$\pm$0.18 &  94.38$\pm$0.04 &0.023$\pm$0.001 &0.72$\pm$0.00 &  68.06$\pm$2.23 & 45.63$\pm$2.45 &  0.065$\pm$0.002 &  0.57$\pm$0.01 & 60.62$\pm$0.28 & 45.42$\pm$0.26 & 0.08$\pm$0.00& 0.55$\pm$0.01 \\

\textbf{LUR} & 	88.87$\pm$0.32 & 83.24$\pm$0.18 & 0.060$\pm$0.040 & 0.72$\pm$0.03	&	94.19$\pm$0.34 & 95.16$\pm$0.29 & 0.020$\pm$0.009 & 0.72$\pm$0.06	&	68.66$\pm$1.33 & 55.82$\pm$2.01 & 0.080$\pm$0.011 & 0.46$\pm$0.12	&	60.52$\pm$0.13 & 45.40$\pm$0.07 & 0.08$\pm$0.00 & 0.56$\pm$0.00 \\ \midrule

\textbf{BBB--LL} & 88.67$\pm$0.36 &  83.13$\pm$0.26 &  0.032$\pm$0.005 & 0.73$\pm$0.02&    93.16$\pm$0.94 &94.28$\pm$0.87 &0.025$\pm$0.007 & 0.65$\pm$0.08  & 71.64$\pm$3.13 & 56.14$\pm$4.37 & 0.068$\pm$0.003 &  0.50$\pm$0.13 & 58.81$\pm$0.94 & 44.70$\pm$1.41 & 0.12$\pm$0.01 & 0.52$\pm$0.03\\

\textbf{BBB--LUR} & 88.47$\pm$0.35 & 82.82$\pm$0.39 & 0.030$\pm$0.001 & 0.70$\pm$0.03 & 93.50$\pm$0.42 & 94.62$\pm$0.29 & 0.030$\pm$0.010 & 0.66$\pm$0.11 & 65.88$\pm$4.78 & 41.47$\pm$2.83 & 0.111$\pm$0.012 & 0.66$\pm$0.08 & 60.10$\pm$0.59 & 44.00$\pm$0.73 & 0.13$\pm$0.06 & 0.51$\pm$0.09 \\

\midrule

\textbf{PE--LL} & 88.64$\pm$0.32 &83.37$\pm$0.02 &  0.030$\pm$0.003 & 0.74$\pm$0.01 & 94.02$\pm$2.24  &   95.02$\pm$0.02 &   0.022$\pm$0.005 & 0.67$\pm$0.03 & 68.66$\pm$2.11 & 59.28$\pm$1.38 & 0.077$\pm$0.008 &  0.36$\pm$0.09 & 61.09$\pm$0.19 & 45.74$\pm$0.10 & 0.09$\pm$0.01 & 0.56$\pm$0.01 \\

\textbf{PE--LUR} & 88.67$\pm$0.25 & 83.35$\pm$0.23 & 0.050$\pm$0.020 & 0.72$\pm$0.06 & 94.02$\pm$0.00 & 95.01$\pm$0.01 & 0.020$\pm$0.010 & 0.69$\pm$0.05 & 68.36$\pm$6.22 & 57.84$\pm$1.51 & 0.090$\pm$0.020 & 0.52$\pm$0.07 & 60.73$\pm$0.26 & 45.68$\pm$0.25 & 0.08$\pm$0.00 & 0.56$\pm$0.01 \\ 

\bottomrule
\end{tabular}
}
\label{tab:ll_lur_perf}
\end{table*}

\begin{table*}[!ht]
\centering
\caption{Average LUR and last layer OOD results for the top-performing hyperparameter configurations for the AIDE, Brain4Cars, ROAD, and NuScenes datasets. ROC-AUC=Receiver Operating Characteristic - Area Under the Curve, PR=Precision-Recall - Area Under the Curve, FPR95=False Positive Rate at 95\% True Positive Rate. } 
\vspace{-0.1cm}
\resizebox{\textwidth}{!}{
\begin{tabular}{l|c|c|c|c|c|c||c|c|c|c|c|c||c|c|c||c|c|c|c|c|c}
\toprule
\multicolumn{1}{c||}{} & \multicolumn{6}{c||}{\textbf{AIDE}} & \multicolumn{6}{c||}{\textbf{B4C}} & \multicolumn{3}{c||}{\textbf{ROAD}} & \multicolumn{6}{c}{\textbf{NuScenes}} \\
\cmidrule(lr){2-7}\cmidrule(lr){8-13}\cmidrule(lr){14-16}\cmidrule(lr){17-22}
\multicolumn{1}{c||}{} & \multicolumn{3}{c|}{\textbf{OOD min}} & \multicolumn{3}{c||}{\textbf{OOD max}} & \multicolumn{3}{c|}{\textbf{OOD min}} & \multicolumn{3}{c||}{\textbf{OOD max}} & \multicolumn{3}{c||}{\textbf{OOD min}} & \multicolumn{3}{c|}{\textbf{OOD min}} & \multicolumn{3}{c}{\textbf{OOD max}} \\
\multicolumn{1}{c||}{\textbf{}} & 
\begin{tabular}{@{}c@{}}\textbf{ROC-}\\ \textbf{AUC} ($\uparrow$)\end{tabular} & 
\begin{tabular}{@{}c@{}}\textbf{PR-}\\\textbf{AUC} ($\uparrow$)\end{tabular} & 
\begin{tabular}{@{}c@{}}\textbf{FPR95}\\($\downarrow$)\end{tabular} & 
\begin{tabular}{@{}c@{}}\textbf{ROC-}\\\textbf{AUC} ($\uparrow$)\end{tabular} & 
\begin{tabular}{@{}c@{}}\textbf{PR-}\\\textbf{AUC} ($\uparrow$)\end{tabular} & 
\begin{tabular}{@{}c@{}}\textbf{FPR95}\\($\downarrow$)\end{tabular} & 
\begin{tabular}{@{}c@{}}\textbf{ROC-}\\\textbf{AUC} ($\uparrow$)\end{tabular} & 
\begin{tabular}{@{}c@{}}\textbf{PR-}\\\textbf{AUC} ($\uparrow$)\end{tabular} & 
\begin{tabular}{@{}c@{}}\textbf{FPR95}\\($\downarrow$)\end{tabular} & 
\begin{tabular}{@{}c@{}}\textbf{ROC-}\\\textbf{AUC} ($\uparrow$)\end{tabular} & 
\begin{tabular}{@{}c@{}}\textbf{PR-}\\\textbf{AUC} ($\uparrow$)\end{tabular} & 
\begin{tabular}{@{}c@{}}\textbf{FPR95}\\($\downarrow$)\end{tabular} & 
\begin{tabular}{@{}c@{}}\textbf{ROC-}\\\textbf{AUC} ($\uparrow$)\end{tabular} & 
\begin{tabular}{@{}c@{}}\textbf{PR-}\\\textbf{AUC} ($\uparrow$)\end{tabular} & 
\begin{tabular}{@{}c@{}}\textbf{FPR95}\\($\downarrow$)\end{tabular} & 
\begin{tabular}{@{}c@{}}\textbf{ROC-}\\\textbf{AUC} ($\uparrow$)\end{tabular} & 
\begin{tabular}{@{}c@{}}\textbf{PR-}\\\textbf{AUC} ($\uparrow$)\end{tabular} & 
\begin{tabular}{@{}c@{}}\textbf{FPR95}\\($\downarrow$)\end{tabular} & 
\begin{tabular}{@{}c@{}}\textbf{ROC-}\\\textbf{AUC} ($\uparrow$)\end{tabular} & 
\begin{tabular}{@{}c@{}}\textbf{PR-}\\\textbf{AUC} ($\uparrow$)\end{tabular} & 
\begin{tabular}{@{}c@{}}\textbf{FPR95}\\($\downarrow$)\end{tabular} \\

\midrule

\textbf{SE} & 0.51$\pm$0.01 & 0.17$\pm$0.00& 0.86$\pm$0.02	&	0.81$\pm$0.00& 0.93$\pm$0.00& 0.41$\pm$0.01	&	0.29$\pm$0.00& 0.29$\pm$0.00& 1.00$\pm$0.00	&	0.32$\pm$0.02 & 0.69$\pm$0.01 & 0.99$\pm$0.01	&	0.17$\pm$0.01 & 0.24$\pm$0.00& 1.00$\pm$0.00 & 0.53$\pm$0.01 & 0.20$\pm$0.01 & 0.93$\pm$0.01 & 0.65$\pm$0.01 & 0.79$\pm$0.00& 0.84$\pm$0.01 \\

  \textbf{LUR} &  1.00$\pm$0.00&  0.97$\pm$0.01 &  0.00$\pm$0.00	& 1.00$\pm$0.00&  1.00$\pm$0.00& 0.01$\pm$0.01	&	0.97$\pm$0.00&   0.91$\pm$0.01 &  0.05$\pm$0.01	& 	 0.99$\pm$0.00&  1.00$\pm$0.00& 0.04$\pm$0.02	& 1.00$\pm$0.00&1.00$\pm$0.00&0.00$\pm$0.00	& 	0.99$\pm$0.00 &  0.93$\pm$0.01 &  0.03$\pm$0.00	& 	0.99$\pm$0.01 &  0.99$\pm$0.00&  0.04$\pm$0.01\\ \midrule

\textbf{BBB--LL} &  0.95$\pm$0.02 & 0.72$\pm$0.08 &  0.10$\pm$0.05	& 0.96$\pm$0.01 &  0.99$\pm$0.01 & 0.22$\pm$0.06	&  0.84$\pm$0.03 &  0.70$\pm$0.04 & 0.48$\pm$0.21	&  0.80$\pm$0.02 &  0.92$\pm$0.01 &  0.69$\pm$0.10 & 0.60$\pm$0.10 & 0.49$\pm$0.13 & 0.85$\pm$0.06 & 0.76$\pm$0.01 & 0.53$\pm$0.03 & 0.78$\pm$0.04 & 0.78$\pm$0.02 & 0.89$\pm$0.01 & 0.72$\pm$0.04  \\ 

\textbf{BBB--LUR} & 0.89$\pm$0.01 & 0.45$\pm$0.01 & 0.15$\pm$0.01  &  1.00$\pm$0.00 & 1.00$\pm$0.00 & 0.00$\pm$0.00 & 0.99$\pm$0.00 & 0.98$\pm$0.01 & 0.02$\pm$0.01 & 1.00$\pm$0.00 & 1.00$\pm$0.00 & 0.01$\pm$0.02 & 1.00$\pm$0.00 & 1.00$\pm$0.00 & 0.00$\pm$0.00 & 1.00$\pm$0.00 & 0.98$\pm$0.01 & 0.02$\pm$0.00 & 1.00$\pm$0.00 & 1.00$\pm$0.00 & 0.02$\pm$0.01 \\ \midrule

\textbf{PE--LL} &  0.55$\pm$0.01 & 0.52$\pm$0.01 & 0.83$\pm$0.01	& 0.56$\pm$0.01 & 0.53$\pm$0.01 & 0.72$\pm$0.02 &	0.52$\pm$0.01 & 0.52$\pm$0.01 & 0.91$\pm$0.01 &	0.53$\pm$0.01 & 0.53$\pm$0.00 & 0.94$\pm$0.02 &	0.53$\pm$0.01 &  0.52$\pm$0.01 & 0.92$\pm$0.01 & 0.52$\pm$0.01 & 0.52$\pm$0.00 & 0.93$\pm$0.00 & 0.52$\pm$0.00& 0.52$\pm$0.00& 0.93$\pm$0.01 \\
\textbf{PE--LUR} &  1.00$\pm$0.00 & 1.00$\pm$0.00 & 0.00$\pm$0.00 & 1.00$\pm$0.00 & 1.00$\pm$0.00 & 0.00$\pm$0.01 & 0.98$\pm$0.01 & 0.95$\pm$0.03 & 0.03$\pm$0.02 &  0.99$\pm$0.01 & 1.00$\pm$0.00 & 0.02$\pm$0.01 & 1.00$\pm$0.00 & 1.00$\pm$0.00 & 0.00$\pm$0.00 & 0.99$\pm$0.01 & 0.95$\pm$0.02 & 0.02$\pm$0.01 & 1.00$\pm$0.00 & 1.00$\pm$0.00 & 0.01$\pm$0.01   \\

\bottomrule
\end{tabular}
}
\label{tab:ll_vs_lur}
\end{table*}

\newpage
\section{Last layer vs. Latent uncertainty representation} \label{app:ll_vs_lur}
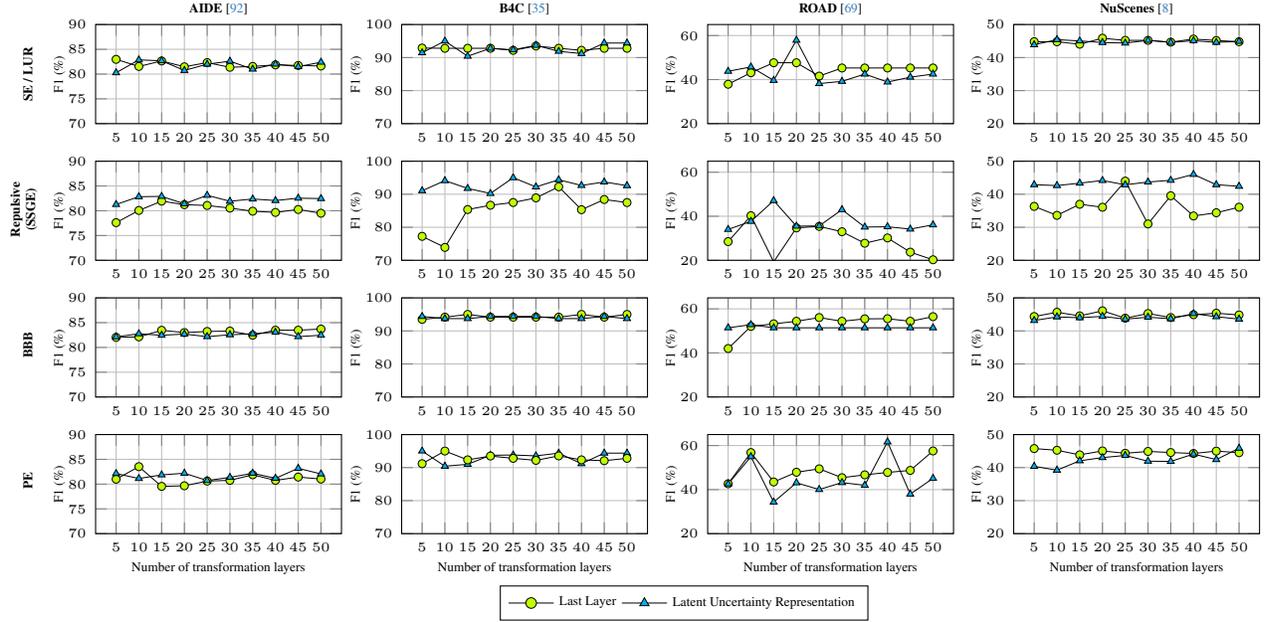
\begin{figure*}[h!]
 \centering
\begin{tikzpicture}
   
\begin{groupplot}[
    group style={
        group size=4 by 4,
        horizontal sep=0.8cm,
        vertical sep=0.50cm,
    },
    width=4.85cm,
    height=2.9cm,
    grid=major,
    xlabel={Number of transformation layers},
    xtick={5,10,15,20,25,30,35,40,45,50},
    tick label style={font=\tiny},
    label style={font=\tiny},
    y tick label style={
        /pgf/number format/fixed,
        /pgf/number format/precision=0,
        /pgf/number format/fixed zerofill=true, 
    },
    tick label style={font=\fontsize{4}{4}\selectfont},
    ylabel style={yshift=-0.20cm},  
    xlabel style={yshift=0.1cm},  
]
   
    
\nextgroupplot[
    title={\tiny{\textbf{AIDE} \citep{yang2023aide}}},
    ylabel={\tiny{F1 (\%)}},
          title style={yshift=-0.2cm},
           xlabel=\empty,
    ymin=70,
    ymax=90 
]

            \addplot[mark=*, thin, line width=0.25pt, mark options={fill=lime}, mark size=1.5pt] coordinates {
          (5,82.97)
   (10,81.54)
   (15,82.65)
   (20,81.47)
   (25,82.32)
   (30,81.37)
   (35,81.54)
   (40,81.86)
   (45,81.74)
   (50,81.63)
            };

            \addplot[mark=triangle*, thin, line width=0.25pt, mark options={fill=cyan}, mark size=1.5pt] coordinates {
    (5,80.27)
   (10,82.87)
   (15,82.68)
   (20,80.73)
   (25,82.00)
(30,82.6)
(35,81.0)
(40,82.0)
(45,81.5)
(50,82.4)

            };

\nextgroupplot[
    title={\tiny{\textbf{B4C} \citep{jain2015car}}},
    ylabel={\tiny{F1 (\%)}},
          title style={yshift=-0.2cm},
          xlabel=\empty,
    ymin=70,
    ymax=100 
]

            \addplot[mark=*, thin, line width=0.25pt, mark options={fill=lime}, mark size=1.5pt] coordinates {
    (5,92.88)
   (10,92.81)
   (15,92.81)
   (20,92.81)
   (25,92.19)
   (30,93.50)
   (35,92.81)
   (40,92.19)
   (45,92.81)
   (50,92.81)
            }; 

            \addplot[mark=triangle*, thin, line width=0.25pt, mark options={fill=cyan}, mark size=1.5pt] coordinates {
    (5,91.452)
   (10,95.014)
   (15,90.41)
   (20,92.81)
   (25,92.34)
      (30,93.7)
(35,91.9)
(40,91.2)
(45,94.4)
(50,94.4)
            };

            \nextgroupplot[
    title={\tiny{\textbf{ROAD} \citep{singh2022road}}},
    ylabel={\tiny{F1 (\%)}},
          title style={yshift=-0.2cm},
           xlabel=\empty,
    ymin=20,
    ymax=65 
]
  
            \addplot[mark=*, thin, line width=0.25pt, mark options={fill=lime}, mark size=1.5pt] coordinates {
   (5,37.89)
   (10,43.13)
   (15,47.65)
   (20,47.65)
   (25,41.58)
   (30,45.27)
   (35,45.27)
   (40,45.27)
   (45,45.27)
   (50,45.27)
            };

                \addplot[mark=triangle*, thin, line width=0.25pt, mark options={fill=cyan}, mark size=1.5pt] coordinates {
    (5,43.78)
   (10,45.74)
   (15,39.63)
   (20,57.89)
   (25,38.18)
   (30,39.2)
(35,42.5)
(40,38.9)
(45,41.1)
(50,42.5)

            };
            
            \nextgroupplot[
    title={\tiny{\textbf{NuScenes} \citep{caesar2020nuscenes}}},
    ylabel={\tiny{F1 (\%)}},
          title style={yshift=-0.2cm},
     xlabel=\empty,
    ymin=20,
    ymax=50 
]
  
            \addplot[mark=*, thin, line width=0.25pt, mark options={fill=lime}, mark size=1.5pt] coordinates {
          (5,44.81)
  (10,44.789)
(15,44.047)
(20,45.852)
(25,45.208)
(30,45.208)
(35,44.685)
(40,45.604)
(45,45.163)
(50,44.805)
            };

            \addplot[mark=triangle*, thin, line width=0.25pt, mark options={fill=cyan}, mark size=1.5pt] coordinates {
    (5,43.88)
   (10,45.42)
   (15,45.04)
   (20,44.55)
   (25,44.39)
   (30,45.2)
(35,44.6)
(40,45.1)
(45,44.6)
(50,44.9)
            };

\nextgroupplot[
    ylabel={\tiny{F1 (\%)}},
          title style={yshift=-0.2cm},
     xlabel=\empty,
    ymin=70,
    ymax=90 
]
\addplot[mark=*, thin, line width=0.25pt, mark options={fill=lime}, mark size=1.5pt] coordinates {
        (5,77.601)
(10,80.061)
(15,81.944)
(20,81.23)
(25,81.066)
(30,80.539)
(35,79.91)
(40,79.662)
(45,80.256)
(50,79.505)
            };

            \addplot[mark=triangle*, thin, line width=0.25pt, mark options={fill=cyan}, mark size=1.5pt] coordinates {

(5, 81.271)
(10,82.834)
(15,82.90)
(20,81.406)
(25,83.114)
(30,81.92)
(35,82.374)
(40,82.061)
(45,82.539)
(50,82.459)
            };

\nextgroupplot[
    ylabel={\tiny{F1 (\%)}},
          title style={yshift=-0.2cm},
           xlabel=\empty,
    ymin=70,
    ymax=100 
]

                    \addplot[mark=*, thin, line width=0.25pt, mark options={fill=lime}, mark size=1.5pt] coordinates {
   (5,77.252)
(10,73.905)
(15,85.367)
(20,86.671)
(25,87.497)
(30,88.897)
(35,92.26)
(40,85.324)
(45,88.406)
(50,87.482)
            }; 

    \addplot[mark=triangle*, thin, line width=0.25pt, mark options={fill=cyan}, mark size=1.5pt] coordinates {
(5,91.063)
(10,94.111)
(15,91.763)
(20,90.237)
(25,95.003)
(30,92.217)
(35,94.382)
(40,92.658)
(45,93.715)
(50,92.595)
            };

            \nextgroupplot[
    ylabel={\tiny{F1 (\%)}},
          title style={yshift=-0.2cm},
           xlabel=\empty,
     xlabel=\empty,
    ymin=20,
    ymax=65 
]
  
                    \addplot[mark=*, thin, line width=0.25pt, mark options={fill=lime}, mark size=1.5pt] coordinates {
(5,28.497)
(10,40.246)
(15,19.295)
(20,34.72)
(25,35.385)
(30,32.975)
(35,27.757)
(40,30.157)
(45,23.701)
(50,20.242)
            }; 

    \addplot[mark=triangle*, thin, line width=0.25pt, mark options={fill=cyan}, mark size=1.5pt] coordinates {
(5,34.038)
(10,37.67)
(15,47.111)
(20,35.5)
(25,35.613)
(30,43.004)
(35,35.093)
(40,35.234)
(45,34.179)
(50,36.096)
            };
            \nextgroupplot[
    ylabel={\tiny{F1 (\%)}},
          title style={yshift=-0.2cm},
           xlabel=\empty,
    ymin=20,
    ymax=50 
]

                    \addplot[mark=*, thin, line width=0.25pt, mark options={fill=lime}, mark size=1.5pt] coordinates {
(5,36.325)
(10,33.605)
(15,36.97)
(20,36.066)
(25,43.977)
(30,31.01)
(35,39.539)
(40,33.427)
(45,34.39)
(50,36.074)
            }; 

    \addplot[mark=triangle*, thin, line width=0.25pt, mark options={fill=cyan}, mark size=1.5pt] coordinates {
(5,42.887)
(10,42.622)
(15,43.356)
(20,44.165)
(25,42.848)
(30,43.703)
(35,44.221)
(40,46.01)
(45,42.881)
(50,42.408)
            };
            
\nextgroupplot[
    ylabel={\tiny{F1 (\%)}},
          title style={yshift=-0.2cm},
     xlabel=\empty,
    ymin=70,
    ymax=90 
]

\addplot[mark=*, thin, line width=0.25pt, mark options={fill=lime}, mark size=1.5pt] coordinates {
  (5,82.04)
   (10,82.13)
   (15,83.46 )
   (20,83.00)
   (25,83.22)
   (30,83.31)
   (35,82.47)
   (40,83.48)
   (45,83.48)
   (50,83.70)
};

    \addplot[mark=triangle*, thin, line width=0.25pt, mark options={fill=cyan}, mark size=1.5pt] coordinates {
(5,82.133)
(10,82.754)
(15,82.459)
(20,82.698)
(25,82.181)
(30,82.554)
(35,82.754)
(40,83.063)
(45,82.181)
(50,82.459)
            };

\nextgroupplot[
    ylabel={\tiny{F1 (\%)}},
          title style={yshift=-0.2cm},
           xlabel=\empty,
    ymin=70,
    ymax=100 
]
  \addplot[mark=*, thin, line width=0.25pt, mark options={fill=lime}, mark size=1.5pt] coordinates {
   (5,93.50)
   (10,94.17)
   (15,95.00)
   (20,94.17)
   (25,94.17)
   (30,94.17)
   (35,94.17)
   (40,95.00)
   (45,94.17)
   (50,95.00)
};

    \addplot[mark=triangle*, thin, line width=0.25pt, mark options={fill=cyan}, mark size=1.5pt] coordinates {
(5,94.382)
(10,93.733)
(15,93.733)
(20,94.382)
(25,94.382)
(30,94.382)
(35,93.733)
(40,93.733)
(45,94.382)
(50,93.733)
};
            \nextgroupplot[
    ylabel={\tiny{F1 (\%)}},
          title style={yshift=-0.2cm},
           xlabel=\empty,
     xlabel=\empty,
    ymin=20,
    ymax=65 
]
    \addplot[mark=*, thin, line width=0.25pt, mark options={fill=lime}, mark size=1.5pt] coordinates {
          (5,42.00)
   (10,52.06)
   (15,53.21)
   (20,54.38)
   (25,56.06)
   (30,54.38)
   (35,55.47)
   (40,55.55)
   (45,54.38)
   (50,56.47)
            };

    \addplot[mark=triangle*, thin, line width=0.25pt, mark options={fill=cyan}, mark size=1.5pt] coordinates {
(5,51.41)
(10,52.828)
(15,51.41)
(20,51.41)
(25,51.41)
(30,51.41)
(35,51.41)
(40,51.41)
(45,51.41)
(50,51.41)
};

            \nextgroupplot[
    ylabel={\tiny{F1 (\%)}},
          title style={yshift=-0.2cm},
           xlabel=\empty,
    ymin=20,
    ymax=50 
]
    \addplot[mark=*, thin, line width=0.25pt, mark options={fill=lime}, mark size=1.5pt] coordinates {
       (5,44.358)
(10,45.649)
(15,44.555)
(20,46.089)
(25,43.858)
(30,45.267)
(35,44.055)
(40,44.919)
(45,45.366)
(50,44.836)
            };
            
\addplot[mark=triangle*, thin, line width=0.25pt, mark options={fill=cyan}, mark size=1.5pt] coordinates {
(5,43.165)
(10,44.238)
(15,43.99)
(20,44.405)
(25,43.57)
(30,44.157)
(35,43.686)
(40,45.227)
(45,44.306)
(50,43.578)
};

\nextgroupplot[
    ylabel={\tiny{F1 (\%)}},
          title style={yshift=-0.2cm},
    ymin=70,
    ymax=90 
]

          \addplot[mark=*, thin, line width=0.25pt, mark options={fill=lime}, mark size=1.5pt] coordinates {
            (5,80.99)
   (10,83.52)
   (15,79.54)
   (20,79.66)
   (25,80.61)
   (30,80.76)
   (35,81.88)
   (40,80.75)
   (45,81.42)
   (50,81.02)
            };
   \addplot[mark=triangle*, thin, line width=0.25pt, mark options={fill=cyan}, mark size=1.5pt] coordinates {
(5,82.132)
(10,81.129)
(15,81.858)
(20,82.193)
(25,80.723)
(30,81.356)
(35,82.193)
(40,81.168)
(45,83.185)
(50,82.054)};

\nextgroupplot[
    ylabel={\tiny{F1 (\%)}},
          title style={yshift=-0.2cm},
    ymin=70,
    ymax=100 
]

  \addplot[mark=triangle*, thin, line width=0.25pt, mark options={fill=cyan}, mark size=1.5pt] coordinates {
         (5,95.03)
   (10,90.42)
   (15,91.00)
   (20,93.73)
   (25,93.81)
   (30,93.55)
   (35,94.38)
   (40,91.18)
   (45,94.38)
   (50,94.38)
            };
    
  \addplot[mark=*, thin, line width=0.25pt, mark options={fill=lime}, mark size=1.5pt] coordinates {
            (5,91.181)
(10,95.015)
(15,92.343)
(20,93.496)
(25,92.806)
(30,92.191)
(35,93.547)
(40,92.343)
(45,92.095)
(50,92.806) };
            \nextgroupplot[
    ylabel={\tiny{F1 (\%)}},
          title style={yshift=-0.2cm},
    ymin=20,
    ymax=65 
]
  \addplot[mark=*, thin, line width=0.25pt, mark options={fill=lime}, mark size=1.5pt] coordinates {
     (5,42.66)
   (10,56.82)
   (15,43.47)
   (20,47.89)
   (25,49.43)
   (30,45.41)
   (35,46.7)
   (40,47.80)
   (45,48.72)
   (50,57.55)
            };
 \addplot[mark=triangle*, thin, line width=0.25pt, mark options={fill=cyan}, mark size=1.5pt] coordinates {
     (5,42.564)
(10,54.938)
(15,34.341)
(20,43.091)
(25,40.043)
(30,43.181)
(35,41.955)
(40,61.641)
(45,37.968)
(50,45.104) };

            \nextgroupplot[
    ylabel={\tiny{F1 (\%)}},
          title style={yshift=-0.2cm},
    ymin=20,
    ymax=50 
]

          \addplot[mark=*, thin, line width=0.25pt, mark options={fill=lime}, mark size=1.5pt] coordinates {
        (5,45.758)
(10,45.251)
(15,43.86)
(20,45.046)
(25,44.333)
(30,44.892)
(35,44.562)
(40,44.272)
(45,44.992)
(50,44.515)
            }; 
 \addplot[mark=triangle*, thin, line width=0.25pt, mark options={fill=cyan}, mark size=1.5pt] coordinates {
(5,40.387)
(10,39.217)
(15,42.059)
(20,43.03)
(25,43.673)
(30,41.945)
(35,41.871)
(40,44.092)
(45,42.425)
(50,45.857)};

        \end{groupplot}
          \node[font=\bfseries\tiny, rotate=90] at ($(group c1r1.west)+(-0.9cm,-0.0)$) {SE / LUR};
          \node[font=\bfseries\tiny, rotate=90] at ($(group c1r1.west)+(-1.05cm,-1.8)$) {Repulsive};
          \node[font=\bfseries\tiny, rotate=90] at ($(group c1r1.west)+(-0.85cm,-1.8)$) {(SSGE)};
        \node[font=\bfseries\tiny, rotate=90] at ($(group c1r1.west)+(-0.9cm,-3.6)$) {BBB};
        \node[font=\bfseries\tiny, rotate=90] at ($(group c1r1.west)+(-0.9cm,-5.4)$) {PE};
    \end{tikzpicture}
    
\begin{tikzpicture}
\hspace{0.75cm}
\begin{axis}[
    hide axis,
    xmin=0, xmax=1, ymin=0, ymax=1,
    legend style={at={(0.5,0.5)}, anchor=center, legend columns=4},
    legend entries={\tiny{Last Layer},\tiny{Latent Uncertainty Representation}}
]

\addlegendimage{mark=*, thin, line width=0.25pt, mark options={fill=lime}, mark size=2.0pt}
\addlegendimage{mark=triangle*, thin, line width=0.25pt, mark options={fill=cyan}, mark size=2.0pt}
\addlegendimage{mark=square*, thin, line width=0.25pt, mark options={fill=pink}, mark size=2pt}
\addlegendimage{mark=diamond*, thin, line width=0.25pt, mark options={fill=orange}, mark size=2.0pt}

\end{axis}
\end{tikzpicture}
\vspace{-0.05cm}
    \caption{In-distribution performance for the last layer and the equivalent latent uncertainty representation approaches for a single random seed.}
    \label{fig:id_perf_overview_multiple}
    \vspace{0.3cm}
\end{figure*}

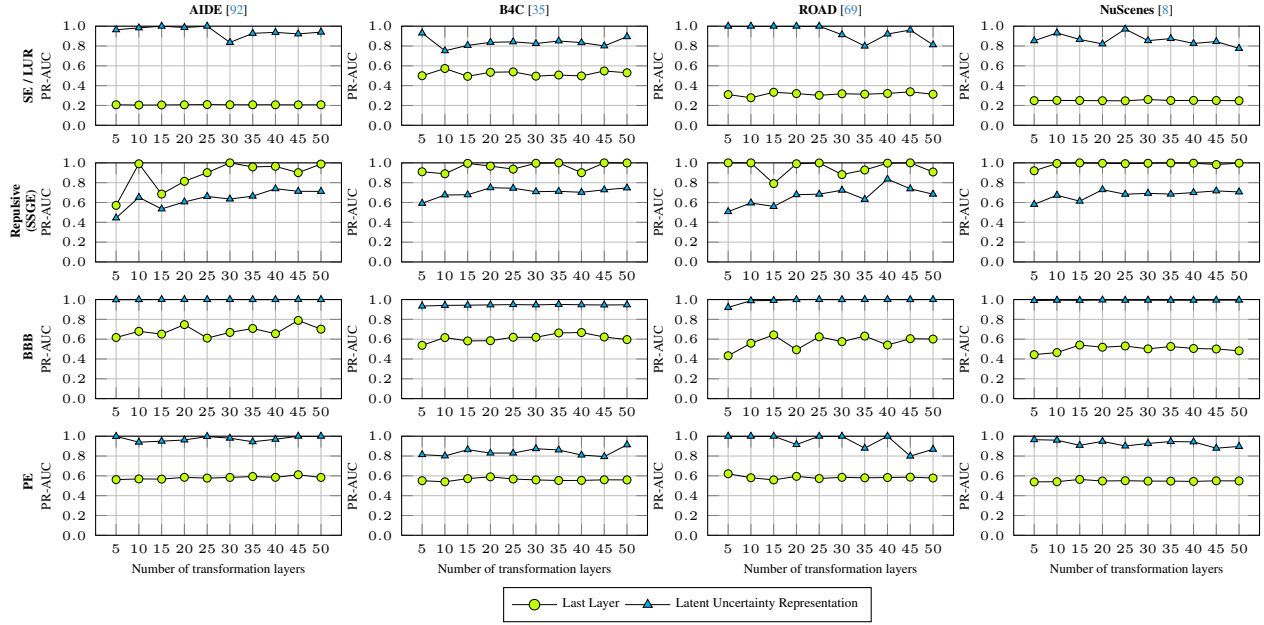
\begin{figure*}[t!]
 \centering
\begin{tikzpicture}
   
\begin{groupplot}[
 group style={
        group size=4 by 4,
        horizontal sep=0.8cm,
        vertical sep=0.50cm,
    },
    width=4.85cm,
    height=2.9cm,
    grid=major,
    xlabel={Number of transformation layers},
    xtick={5,10,15,20,25,30,35,40,45,50},
    tick label style={font=\tiny},
    label style={font=\tiny},
    y tick label style={
        /pgf/number format/fixed,
        /pgf/number format/precision=1,
        /pgf/number format/fixed zerofill=true, 
    },
    tick label style={font=\fontsize{4}{4}\selectfont},
    ylabel style={yshift=-0.09cm},  
    xlabel style={yshift=0.1cm},  
]
   
    
\nextgroupplot[
    title={\tiny{\textbf{AIDE} \citep{yang2023aide}}},
    ylabel={\tiny{PR-AUC}},
          title style={yshift=-0.2cm},
           xlabel=\empty,
    ymin=0.0,
    ymax=1.0 
]

            \addplot[mark=*, thin, line width=0.25pt, mark options={fill=lime}, mark size=1.5pt] coordinates {
  (5,0.207)
(10,0.204)
(15,0.205)
(20,0.207)
(25,0.209)
(30,0.207)
(35,0.207)
(40,0.207)
(45,0.206)
(50,0.207)
            };

            \addplot[mark=triangle*, thin, line width=0.25pt, mark options={fill=cyan}, mark size=1.5pt] coordinates {
(5,0.964)
(10,0.983)
(15,1.)
(20,0.987)
(25,1.)
(30,0.836)
(35,0.928)
(40,0.9377)
(45,0.92248)
(50,0.940)
            };

\nextgroupplot[
    title={\tiny{\textbf{B4C} \citep{jain2015car}}},
    ylabel={\tiny{PR-AUC}},
          title style={yshift=-0.2cm},
          xlabel=\empty,
    ymin=0.0,
    ymax=1.00 
]

            \addplot[mark=*, thin, line width=0.25pt, mark options={fill=lime}, mark size=1.5pt] coordinates {
   (5,0.5)
(10,0.573)
(15,0.494)
(20,0.535)
(25,0.539)
(30,0.496)
(35,0.507)
(40,0.498)
(45,0.548)
(50,0.53)
            }; 

            \addplot[mark=triangle*, thin, line width=0.25pt, mark options={fill=cyan}, mark size=1.5pt] coordinates {
(5,0.931)
(10,0.752)
(15,0.806)
(20,0.837)
(25,0.842)
(30,0.825)
(35,0.851)
(40,0.834)
(45,0.801)
(50,0.894)
            };

            \nextgroupplot[
    title={\tiny{\textbf{ROAD} \citep{singh2022road}}},
    ylabel={\tiny{PR-AUC}},
          title style={yshift=-0.2cm},
           xlabel=\empty,
    ymin=0.0,
    ymax=1.0 
]
  
            \addplot[mark=*, thin, line width=0.25pt, mark options={fill=lime}, mark size=1.5pt] coordinates {
(5,0.31)
(10,0.278)
(15,0.333)
(20,0.32)
(25,0.302)
(30,0.318)
(35,0.314)
(40,0.321)
(45,0.338)
(50,0.313)
            };

                \addplot[mark=triangle*, thin, line width=0.25pt, mark options={fill=cyan}, mark size=1.5pt] coordinates {
(5,1.)
(10,1.)
(15,1.)
(20,1.)
(25,1.)
(30,0.914)
(35,0.799)
(40,0.921)
(45,0.961)
(50,0.812)
            };
            
            \nextgroupplot[
    title={\tiny{\textbf{NuScenes} \citep{caesar2020nuscenes}}},
    ylabel={\tiny{PR-AUC}},
          title style={yshift=-0.2cm},
     xlabel=\empty,
    ymin=0.0,
    ymax=1.0 
]
  
            \addplot[mark=*, thin, line width=0.25pt, mark options={fill=lime}, mark size=1.5pt] coordinates {
 (5,0.249)
(10,0.251)
(15,0.25)
(20,0.248)
(25,0.247)
(30,0.26)
(35,0.25)
(40,0.251)
(45,0.25)
(50,0.248)
            };

            \addplot[mark=triangle*, thin, line width=0.25pt, mark options={fill=cyan}, mark size=1.5pt] coordinates {
(5,0.853)
(10,0.932)
(15,0.866)
(20,0.821)
(25,0.971)
(30,0.854)
(35,0.875)
(40,0.826)
(45,0.845)
(50,0.776)
            };

\nextgroupplot[
    ylabel={\tiny{PR-AUC}},
          title style={yshift=-0.2cm},
     xlabel=\empty,
    ymin=0.0,
    ymax=1.0 
]
\addplot[mark=*, thin, line width=0.25pt, mark options={fill=lime}, mark size=1.5pt] coordinates {
(5,0.571)
(10,0.991)
(15,0.685)
(20,0.814)
(25,0.901)
(30,1.)
(35,0.959)
(40,0.965)
(45,0.901)
(50,0.988)
            };

            \addplot[mark=triangle*, thin, line width=0.25pt, mark options={fill=cyan}, mark size=1.5pt] coordinates {
(5,0.444)
(10,0.652)
(15,0.535)
(20,0.606)
(25,0.661)
(30,0.635)
(35,0.664)
(40,0.739)
(45,0.713)
(50,0.712)
            };

\nextgroupplot[
    ylabel={\tiny{PR-AUC}},
          title style={yshift=-0.2cm},
           xlabel=\empty,
    ymin=0.0,
    ymax=1. 
]

                    \addplot[mark=*, thin, line width=0.25pt, mark options={fill=lime}, mark size=1.5pt] coordinates {
(5,0.91)
(10,0.89)
(15,0.995)
(20,0.967)
(25,0.937)
(30,0.996)
(35,1.)
(40,0.901)
(45,1.)
(50,0.999)
            }; 

    \addplot[mark=triangle*, thin, line width=0.25pt, mark options={fill=cyan}, mark size=1.5pt] coordinates {
(5,0.592)
(10,0.676)
(15,0.678)
(20,0.749)
(25,0.744)
(30,0.709)
(35,0.713)
(40,0.703)
(45,0.728)
(50,0.747)
            };

            \nextgroupplot[
    ylabel={\tiny{PR-AUC}},
          title style={yshift=-0.2cm},
           xlabel=\empty,
     xlabel=\empty,
    ymin=0.0,
    ymax=1.0 
]
  
                    \addplot[mark=*, thin, line width=0.25pt, mark options={fill=lime}, mark size=1.5pt] coordinates {
(5,1.)
(10,1.)
(15,0.791)
(20,0.991)
(25,0.999)
(30,0.882)
(35,0.928)
(40,0.997)
(45,1.)
(50,0.908)
            }; 

    \addplot[mark=triangle*, thin, line width=0.25pt, mark options={fill=cyan}, mark size=1.5pt] coordinates {
(5,0.508)
(10,0.595)
(15,0.56)
(20,0.679)
(25,0.684)
(30,0.724)
(35,0.632)
(40,0.836)
(45,0.738)
(50,0.684)
            };
            \nextgroupplot[
    ylabel={\tiny{PR-AUC}},
          title style={yshift=-0.2cm},
           xlabel=\empty,
    ymin=0.0,
    ymax=1.0 
]

                    \addplot[mark=*, thin, line width=0.25pt, mark options={fill=lime}, mark size=1.5pt] coordinates {
(5,0.92)
(10,0.994)
(15,1.)
(20,0.996)
(25,0.992)
(30,0.996)
(35,0.999)
(40,0.997)
(45,0.983)
(50,0.997)
            }; 

    \addplot[mark=triangle*, thin, line width=0.25pt, mark options={fill=cyan}, mark size=1.5pt] coordinates {
(5,0.58119)
(10,0.67323)
(15,0.6133)
(20,0.72949)
(25,0.68316)
(30,0.69285)
(35,0.68475)
(40,0.70027)
(45,0.71699)
(50,0.70653)
            };
            
\nextgroupplot[
    ylabel={\tiny{PR-AUC}},
          title style={yshift=-0.2cm},
     xlabel=\empty,
    ymin=0.0,
    ymax=1.0 
]

\addplot[mark=*, thin, line width=0.25pt, mark options={fill=lime}, mark size=1.5pt] coordinates {
 (5,0.617)
(10,0.679)
(15,0.651)
(20,0.747)
(25,0.611)
(30,0.668)
(35,0.708)
(40,0.655)
(45,0.789)
(50,0.701)
};     

\addplot[mark=triangle*, thin, line width=0.25pt, mark options={fill=cyan}, mark size=1.5pt] coordinates {
(5,0.998)
(10,0.999)
(15,1.)
(20,1.)
(25,1.)
(30,1.)
(35,0.999)
(40,0.999)
(45,1.)
(50,1.)
};

\nextgroupplot[
    ylabel={\tiny{PR-AUC}},
          title style={yshift=-0.2cm},
           xlabel=\empty,
    ymin=0.0,
    ymax=1.0 
]
  \addplot[mark=*, thin, line width=0.25pt, mark options={fill=lime}, mark size=1.5pt] coordinates {
  (5,0.538)
(10,0.616)
(15,0.582)
(20,0.585)
(25,0.619)
(30,0.619)
(35,0.663)
(40,0.667)
(45,0.622)
(50,0.596)
};

\addplot[mark=triangle*, thin, line width=0.25pt, mark options={fill=cyan}, mark size=1.5pt] coordinates {
(5,0.934)
(10,0.941)
(15,0.943)
(20,0.946)
(25,0.949)
(30,0.946)
(35,0.951)
(40,0.947)
(45,0.945)
(50,0.947)
};

            \nextgroupplot[
    ylabel={\tiny{PR-AUC}},
          title style={yshift=-0.2cm},
           xlabel=\empty,
     xlabel=\empty,
    ymin=0.0,
    ymax=1.0 
]
    \addplot[mark=*, thin, line width=0.25pt, mark options={fill=lime}, mark size=1.5pt] coordinates {
 (5,0.433)
(10,0.559)
(15,0.643)
(20,0.493)
(25,0.624)
(30,0.575)
(35,0.631)
(40,0.541)
(45,0.605)
(50,0.601)
            };
            
\addplot[mark=triangle*, thin, line width=0.25pt, mark options={fill=cyan}, mark size=1.5pt] coordinates {
            (5,0.921)
(10,0.988)
(15,0.991)
(20,0.999)
(25,1.)
(30,0.999)
(35,1.)
(40,1.)
(45,0.999)
(50,1.)};
            
            \nextgroupplot[
    ylabel={\tiny{PR-AUC}},
          title style={yshift=-0.2cm},
           xlabel=\empty,
    ymin=0.0,
    ymax=1.0 
]
    \addplot[mark=*, thin, line width=0.25pt, mark options={fill=lime}, mark size=1.5pt] coordinates {
  (5,0.444)
(10,0.465)
(15,0.541)
(20,0.519)
(25,0.531)
(30,0.502)
(35,0.526)
(40,0.506)
(45,0.501)
(50,0.483)
            };
\addplot[mark=triangle*, thin, line width=0.25pt, mark options={fill=cyan}, mark size=1.5pt] coordinates {
            (5,0.99)
(10,0.993)
(15,0.992)
(20,0.995)
(25,0.993)
(30,0.993)
(35,0.993)
(40,0.993)
(45,0.994)
(50,0.994)
};
            
\nextgroupplot[
    ylabel={\tiny{PR-AUC}},
          title style={yshift=-0.2cm},
    ymin=0.00,
    ymax=1.00 
]

          \addplot[mark=*, thin, line width=0.25pt, mark options={fill=lime}, mark size=1.5pt] coordinates {
    (5,0.562)
(10,0.569)
(15,0.567)
(20,0.585)
(25,0.577)
(30,0.584)
(35,0.593)
(40,0.585)
(45,0.611)
(50,0.584)
            };

\addplot[mark=triangle*, thin, line width=0.25pt, mark options={fill=cyan}, mark size=1.5pt] coordinates {
            (5,0.998)
(10,0.94)
(15,0.95)
(20,0.962)
(25,0.997)
(30,0.98)
(35,0.944)
(40,0.969)
(45,1.)
(50,1.) };

\nextgroupplot[
    ylabel={\tiny{PR-AUC}},
          title style={yshift=-0.2cm},
    ymin=0.00,
    ymax=1.00 
]

 \addplot[mark=*, thin, line width=0.25pt, mark options={fill=lime}, mark size=1.5pt] coordinates {
      (5,0.551)
(10,0.54)
(15,0.572)
(20,0.59)
(25,0.568)
(30,0.559)
(35,0.553)
(40,0.554)
(45,0.56)
(50,0.559)
            };

\addplot[mark=triangle*, thin, line width=0.25pt, mark options={fill=cyan}, mark size=1.5pt] coordinates {
            (5,0.814)
(10,0.801)
(15,0.864)
(20,0.83)
(25,0.83)
(30,0.874)
(35,0.861)
(40,0.81)
(45,0.794)
(50,0.913) };

            \nextgroupplot[
    ylabel={\tiny{PR-AUC}},
          title style={yshift=-0.2cm},
    ymin=0.0,
    ymax=1.0 
]
  \addplot[mark=*, thin, line width=0.25pt, mark options={fill=lime}, mark size=1.5pt] coordinates {
   (5,0.621)
(10,0.581)
(15,0.559)
(20,0.595)
(25,0.573)
(30,0.585)
(35,0.58)
(40,0.583)
(45,0.587)
(50,0.578)
            };

\addplot[mark=triangle*, thin, line width=0.25pt, mark options={fill=cyan}, mark size=1.5pt] coordinates {
(5,1.)
(10,1.)
(15,1.)
(20,0.916)
(25,1.)
(30,1.)
(35,0.879)
(40,1.)
(45,0.799)
(50,0.868)
};

            \nextgroupplot[
    ylabel={\tiny{PR-AUC}},
          title style={yshift=-0.2cm},
    ymin=0.0,
    ymax=1.0 
]

          \addplot[mark=*, thin, line width=0.25pt, mark options={fill=lime}, mark size=1.5pt] coordinates {
(5,0.539)
(10,0.541)
(15,0.564)
(20,0.548)
(25,0.551)
(30,0.547)
(35,0.547)
(40,0.544)
(45,0.55)
(50,0.549)

            }; 
            
\addplot[mark=triangle*, thin, line width=0.25pt, mark options={fill=cyan}, mark size=1.5pt] coordinates {
(5,0.965)
(10,0.96)
(15,0.908)
(20,0.948)
(25,0.901)
(30,0.927)
(35,0.946)
(40,0.943)
(45,0.879)
(50,0.897)
};

        \end{groupplot}
         \node[font=\bfseries\tiny, rotate=90] at ($(group c1r1.west)+(-0.9cm,-0.0)$) {SE / LUR};
          \node[font=\bfseries\tiny, rotate=90] at ($(group c1r1.west)+(-1.05cm,-1.8)$) {Repulsive};
          \node[font=\bfseries\tiny, rotate=90] at ($(group c1r1.west)+(-0.85cm,-1.8)$) {(SSGE)};
        \node[font=\bfseries\tiny, rotate=90] at ($(group c1r1.west)+(-0.9cm,-3.6)$) {BBB};
        \node[font=\bfseries\tiny, rotate=90] at ($(group c1r1.west)+(-0.9cm,-5.4)$) {PE};
          
    \end{tikzpicture}
    
\begin{tikzpicture}
\hspace{0.75cm}
\begin{axis}[
    hide axis,
    xmin=0, xmax=1, ymin=0, ymax=1,
    legend style={at={(0.5,0.5)}, anchor=center, legend columns=4},
    legend entries={\tiny{Last Layer},\tiny{Latent Uncertainty Representation}}
]

\addlegendimage{mark=*, thin, line width=0.25pt, mark options={fill=lime}, mark size=2.0pt}
\addlegendimage{mark=triangle*, thin, line width=0.25pt, mark options={fill=cyan}, mark size=2.0pt}
\addlegendimage{mark=square*, thin, line width=0.25pt, mark options={fill=pink}, mark size=2pt}
\addlegendimage{mark=diamond*, thin, line width=0.25pt, mark options={fill=orange}, mark size=2.0pt}

\end{axis}
\end{tikzpicture}
    \caption{OOD min performance comparison for the last layer and latent uncertainty representation approaches for a single random seed.}
    \label{fig:OOD_overview_multiple}
\end{figure*}

Tables \ref{tab:ll_lur_perf} and \ref{tab:ll_vs_lur} provide an  overview of the results reported in Figure \ref{fig:translayer_type}. Additionally, we evaluated the influence of the number of transformation layers on both in-distribution performance and OOD detection scenarios. Figures \ref{fig:id_perf_overview_multiple} and \ref{fig:OOD_overview_multiple} visualize the in-distribution classification and OOD min detection performance of the top-performing hyperparameter configurations for a single random seed. Consistent with the findings presented in Figure \ref{fig:translayer_type}, we observe that the performance across different numbers of transformation layers is relatively similar for both the last layer and LUR approaches. For the OOD min scenario, except for the repulsive methods, the LUR approach outperforms the last layer approach, and the number of transformation layers does not appear to have a notable impact.

\end{document}